%% file: main.tex
\documentclass{article} 
\usepackage{style,times}

\input{math_commands.tex}

\usepackage{hyperref}
\hypersetup{
  pdftitle={Daedalus: Bootstrapping Agent Memory from Self-Generated Tasks},
  pdfauthor={Antoine Edy, Max Conti, Victor Xing, Marc-Antoine Allard, Nawfal Benhamdane, Gautier Viaud},
  pdfsubject={Machine Learning},
  pdfkeywords={LLM agents, agent memory, procedural memory, self-generated tasks, self-improvement},
}
\usepackage{url}

\usepackage{amssymb}
\usepackage{pifont}

\usepackage{placeins}
\usepackage{caption}
\usepackage{subcaption}
\usepackage{float}

\usepackage[dvipsnames,svgnames,table]{xcolor}
\usepackage[most]{tcolorbox}

\usepackage{multirow}
\usepackage{multicol}

\usepackage{graphicx}

\usepackage{makecell}

\usepackage{wrapfig}

\usepackage{adjustbox}

\usepackage{inconsolata} 
\usepackage[T1]{fontenc}

\usepackage{array,booktabs}

\usepackage{algorithm, algpseudocode, amsmath}
\algrenewcommand\algorithmicrequire{\textbf{Input:}}
\algrenewcommand\algorithmicensure{\textbf{Output:}}
\newcommand{\agentcall}[1]{\mathsf{#1}}
\newcommand{\envop}[1]{\mathtt{#1}}

\usepackage{xspace}

\usepackage{needspace} 

\usepackage{fontawesome}

\newcommand{\cmark}{\textcolor{OliveGreen}{\ding{51}}}%
\newcommand{\xmark}{\textcolor{BrickRed}{\ding{55}}}%

\newcommand{\up}{\tiny $\uparrow$}
\newcommand{\down}{\tiny $\downarrow$}

\newcommand{\conf}[1]{$_{\pm#1}$}
\newcommand{\confpass}[1]{$_{\pm#1}$}

\tcbset{paperbox/.style={
  enhanced, breakable,
  colback=gray!6, colframe=gray!60!black, boxrule=0.6pt, arc=3pt,
  fonttitle=\small\bfseries,
  left=6pt, right=6pt, top=6pt, bottom=6pt,
  attach boxed title to top left={yshift=-2mm, xshift=4mm},
  boxed title style={colback=gray, colframe=gray!60!black, boxrule=0.6pt, arc=2pt},
  fontupper=\footnotesize, before upper={\setlength{\parindent}{0pt}\raggedright}}}
\newcounter{promptctr}
\definecolor{pGenBack}{HTML}{F1F4F9}\definecolor{pGenFrame}{HTML}{5E7391}   
\definecolor{pJudBack}{HTML}{F8F1F3}\definecolor{pJudFrame}{HTML}{8E6676}   
\definecolor{pMemBack}{HTML}{F1F6F2}\definecolor{pMemFrame}{HTML}{5F8068}   
\definecolor{pSolBack}{HTML}{F9F5EE}\definecolor{pSolFrame}{HTML}{8C7656}   
\tcbset{
  cat/.style 2 args={colback=#1, colframe=#2, boxed title style={colback=#2, colframe=#2}},
  gen/.style={cat={pGenBack}{pGenFrame}}, judge/.style={cat={pJudBack}{pJudFrame}},
  mem/.style={cat={pMemBack}{pMemFrame}}, solver/.style={cat={pSolBack}{pSolFrame}}}
\newtcolorbox{promptbox}[2][]{paperbox, #2, code={\refstepcounter{promptctr}}, title={Prompt \thepromptctr: #1}}
\newtcolorbox{textbox}[2][]{paperbox, #2, title={#1}}
\newcommand{\pline}[1]{#1\par}
\newcommand{\pskip}{\vspace{2pt}}
\newcommand{\phead}[1]{\textbf{#1}\par}
\newcommand{\prole}[1]{{\color{gray!60!black}\textsc{#1}}\par}
\newcommand{\pnote}[1]{{\color{gray!60!black}\textit{[#1]}}\par}
\newcommand{\pelide}{{\color{gray!60!black}[\ldots]}\par}
\newcommand{\pph}[1]{{\color{blue!50!black}\textit{\textless{}#1\textgreater{}}}}
\newcommand{\pitem}[3]{\par\hangindent=#1.1em\hangafter=1\hspace*{\dimexpr#1.1em-1.1em\relax}\makebox[1.1em][l]{#2}#3\par}

\newcommand{\daedalus}{\textsc{Daedalus}\xspace}
\newcommand{\passpowerfive}{pass\textasciicircum{}5\xspace}

\renewcommand{\arraystretch}{1.1} 

\title{\vspace*{-1cm}\daedalus{}: Bootstrapping Agent Memory from Self-Generated Tasks}

\author{%
    Antoine Edy\thanks{Equal Contribution} \xspace \thanks{Corresponding author} \xspace \quad Max Conti$^\star$ \quad Victor Xing$^{\star}$ \\
    \bf Marc-Antoine Allard \quad Nawfal Benhamdane \quad Gautier Viaud \\[2pt]
    {\normalfont Illuin Technology} \\
    {\normalfont\texttt{\{antoine.edy,victor.xing\}@illuin.tech}}
}

\iclrfinalcopy 
\begin{document}

\maketitle
\vspace{-0.5em}
\begin{abstract}
LLM agents often lack the operational knowledge to act reliably in new environments, as they must discover specific tool behaviors or environment conventions on their own.
Without memory of past attempts, they repeat the same mistakes across tasks, leading to more task failures and longer trajectories. To address this, agentic systems typically rely on human-written guidelines or on procedural memory built from training tasks and an oracle verifier, both of which require prior knowledge of the environment.
We present \daedalus, a method for bootstrapping reusable agent memory from self-generated practice without existing tasks or oracle verifiers. \daedalus pairs two agents: an explorer that interacts with the environment to generate challenging yet solvable tasks, and a solver that attempts them. A heuristic is derived from each solver failure and accepted only after the solver repeatedly succeeds with that heuristic in context. These outcomes also provide feedback for the explorer to refine the difficulty of future tasks. The accepted heuristics are then consolidated into a memory bank for test-time use. Across AppWorld, $\tau^2$-bench, and AutomationBench, \daedalus improves mean success rates by up to 15.9 percentage points and \passpowerfive by up to 2.2$\times$ over a no-memory baseline, and is competitive with methods using training tasks, at a lower inference cost than most. We show that performance gains already emerge with a small exploration budget, and that its heuristics also benefit agents from other model families. Our ablations further reveal that solver traces provide the key information needed to derive effective heuristics, while factorizing early discoveries makes exploration more cost-efficient. Beyond memory construction, we find that the tasks generated by \daedalus can serve as a proxy for benchmark tasks when ranking models by performance. We release the code and artifacts, including generation and inference traces, at \href{https://github.com/illuin-tech/daedalus}{\faicon{github} \texttt{\small www.github.com/illuin-tech/daedalus}}.

\end{abstract}

\section{Introduction}
\begin{wrapfigure}[17]{r}{0.44\textwidth} 
    \vspace{-1.7\baselineskip}
    \centering
    \includegraphics[width=\linewidth]{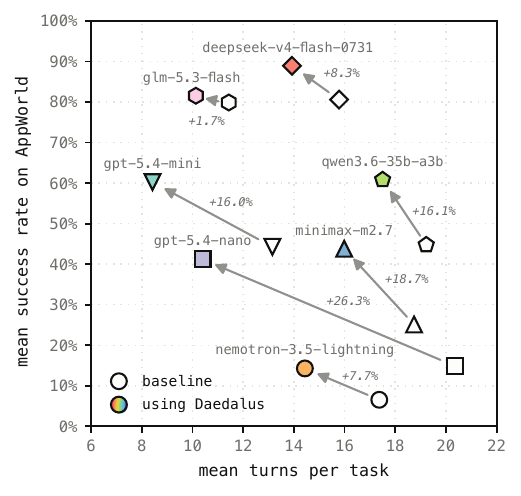}
    \caption{\daedalus{} memory from self-generated tasks improves agent success and efficiency across model families.}
    \label{fig:transferability}
    \vspace{-10\baselineskip}
\end{wrapfigure}

LLM agents are now deployed to perform real-world tasks in a wide range of occupations and industries~\citep{sajadieh2026artificial,anthropic2026economicindex,johnston2026agentic}. As they gain autonomy~\citep{anthropic2026autonomy}, they operate in increasingly complex environments with limited human guidance. Acting reliably in such environments requires practical knowledge that carries over across tasks, a need reflected in the rapid adoption of agent skills~\citep{anthropic2025skills}. Without procedural memory to retain this knowledge, an agent rediscovers its environment from task to task, leading to avoidable errors and longer, costlier executions~\citep{ouyang2026reasoningbank}.

Procedural memory is typically built from past experience: an agent attempts tasks, and lessons drawn from its successes and failures are stored and injected in context at test time~\citep{zhao2024expel,allard2026experiential,zhang2026agentic}, which leaves model weights untouched and applies to both open and closed models. This demands a curated set of training tasks drawn from the target task distribution, often paired with a verifier that scores each attempt. Building this supervision requires prior knowledge of the environment and substantial human effort, as the construction of agentic benchmarks illustrates~\citep{trivedi2024appworld,barres2026taubench}. It is therefore unavailable for most environments, a limitation that is increasingly prevalent as more and more agentic environments are automatically generated without human supervision~\citep{tu2026scaleenv,wang2026agent}. Instead, experience can be gathered from task queries as they arrive at test time~\citep{wang2025agent,ouyang2026reasoningbank}, but the agent then handles its first tasks without memory and improves only after failing on them. In both cases, memory is built only from exposure to the tasks it is meant to support.

In this work, we introduce \daedalus, a training-free method that builds environment-specific memory from self-generated tasks without curated tasks or an oracle verifier. \daedalus{} involves auxiliary agents that explore the environment to create solvable tasks and their success criteria, and a solver agent that attempts these tasks. The solver learns from its failures, as each one is turned into an in-context heuristic that guides its subsequent attempts. The relevance of a heuristic is validated only when it enables the solver to repeatedly succeed on a task it initially failed. Tasks that offer no such learning signal are refined to adjust their difficulty, and these refinements are condensed into task design guidelines for later sessions. Together, these components automate exploration and task creation in an agentic environment, calibrate task difficulty to the solver's capabilities, and produce memory items that are empirically tested.

On AppWorld~\citep{trivedi2024appworld}, $\tau^2$-bench~\citep{barres2026taubench}, and AutomationBench~\citep{shepard2026automationbench}, \daedalus improves success rates by up to 15.9 points over a no-memory baseline and raises \passpowerfive by 1.7 to $2.2\times$, at little to no extra inference cost. It outperforms PREPING~\citep{choi2026preping}, which also builds memory from self-generated tasks, and is competitive with existing methods that use the benchmark training tasks. These gains hold across model families, including when the memory is generated with one family and used by another, and can be achieved with only a small generation budget. Experiments show that validating lessons through the solver's attempts drives these results, whereas memory drawn from exploration alone hurts the agent. We also find that, at this scale, consolidating the bank and injecting it whole at the start of each task outperforms per-turn retrieval. Beyond memory, the generated tasks rank models consistently with the official benchmark test sets, which suggests they can serve as an evaluation proxy when no curated tasks exist. \daedalus{} thus bootstraps agent memory in new environments without human supervision, so that agents no longer discover them through trial and error at test time. We release our code and experimental artifacts, including configurations, generated tasks, heuristic banks, and all agent trajectories, so that our experiments can be reproduced and verified.

\section{Related Work}
\label{sec:related-work}

\paragraph{Memory for LLM agents.}
\label{sec:memory-agents}
LLM-agent memory methods allow agents to accumulate reusable knowledge from past task attempts. Many store it as natural-language insights: ExpeL~\citep{zhao2024expel} contrasts successful and failed trajectories, AutoGuide~\citep{fu2024autoguide} makes such guidelines context-aware, and other systems generate reflections from failed attempts~\citep{allard2026experiential, ouyang2026reasoningbank}. These methods retain an insight based on the outcome of the trajectory it came from, without confirming that it helps the agent overcome the failure; \daedalus{} instead keeps an insight only once it has turned a failure into repeated success. Others build skill libraries, from Voyager's executable code skills~\citep{wang2024voyager} to natural-language skills refined through reward-based optimization~\citep{mi2026skillpro, yang2026skillopt, huang2026skill}, which requires oracle verifiers or weight updates. Memory can also be organized as reusable workflows~\citep{wang2025agent}, evolving playbooks~\citep{zhang2026agentic}, or long-term memory stores~\citep{fang2026trajectory, chhikara2025mem0}. Most of these methods, however, collect experience from a predefined task set, which a new environment does not provide, whereas \daedalus{} generates its own.

\paragraph{Automated agentic task creation.} \label{sec:task-generation}
Automatic exploration is widely used to synthesize agent training data in software, computer-use, and general-application environments~\citep{shi2026taskcraft, ramrakhya2026scaling, dong2026agent}, sometimes with curricula that adapt task difficulty to the agent's progress~\citep{sun2026seagent, xue2026autonomous}. Self-play extends this by training the model as both proposer and solver~\citep{liu2026spade, liu2025spice, lu2026search}, rewarding tasks that are challenging yet verifiable~\citep{yue2026dr, huang2026skill}. In all these works, generated tasks serve to update model weights. Few methods generate tasks to build memory instead: Voyager~\citep{wang2024voyager} proposes tasks through an automatic curriculum, SkillWeaver~\citep{zheng2025skillweaver} practices self-proposed skills on websites, and PREPING~\citep{choi2026preping}, closest to our setting, builds a playbook from self-proposed tasks, updating it from each task's feasibility and completion scores. These methods mainly favor tasks that are feasible or broaden environment coverage, whereas \daedalus{} calibrates tasks to expose failures the agent can learn to overcome.

\section{Building Memory from Self-Generated Tasks}\label{sec:method}

\begin{figure}[t]
    \centering
    \includegraphics[width=1\linewidth]{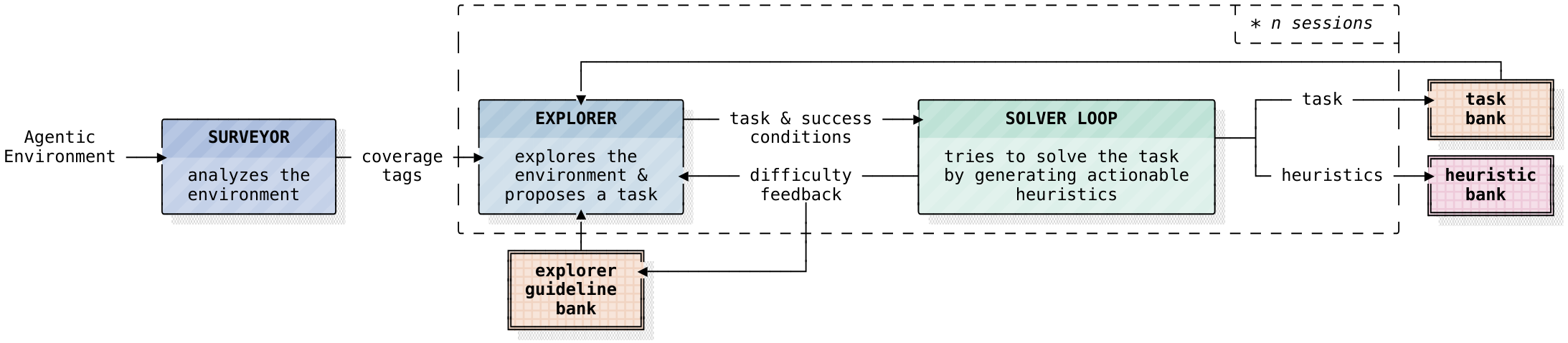}
  \caption{Overview of the \daedalus memory generation pipeline.}
    \label{fig:generation-pipeline}
\end{figure}
\begin{figure}[t]
    \centering
    \includegraphics[width=0.8\linewidth]{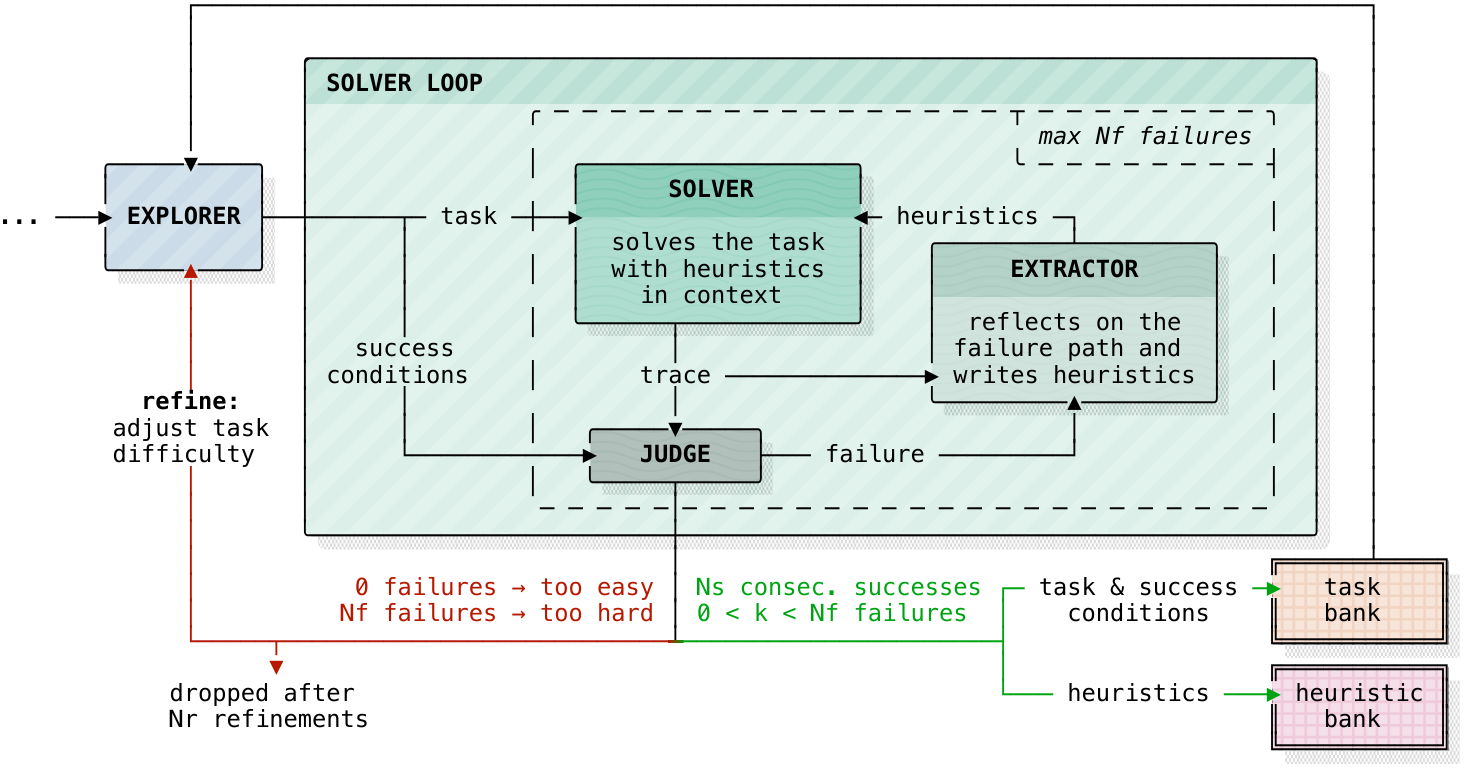}
    \caption{The Solver loop. After each failure, the Extractor writes or revises a heuristic and the Solver retries from a fresh environment state. The heuristic is accepted only after the \emph{Solver} reaches the required streak of successes $N_s$.}
    \label{fig:solver-loop}
\end{figure}

\paragraph{Overview.} We consider a base agent deployed in an agentic environment with no access to training tasks or an oracle verifier, and aim to improve its success rate without updating its weights. To this end, \daedalus{} spends a one-time budget of $n$ practice sessions, bootstrapping a bank of textual heuristics that the base agent receives in context at test time (Figure~\ref{fig:generation-pipeline}). Each session runs two nested loops: in the outer loop, an Explorer agent proposes a task and adjusts its difficulty; in the inner \emph{Solver loop} (Figure~\ref{fig:solver-loop}), a Solver agent attempts the task while an Extractor agent derives a heuristic from its failures. Besides the Solver, generation involves several other LLM agents, including the Explorer and the Extractor, which we call \emph{auxiliary} agents. We describe the Solver loop first, then task calibration and consolidation. Appendix~\ref{apdx pseudo code} formalizes the approach in pseudo-code.

\paragraph{Failure-to-success heuristic validation.} Each task consists of an instruction and explicit success conditions. The Solver attempts the task from a freshly reset environment, and an LLM Judge checks the resulting trajectory against the success conditions; we measure the Judge's agreement with official benchmark verifiers in Section~\ref{sec:judge-validation}. When an attempt fails, the Extractor writes a heuristic from the instruction and the failed trajectory, or revises the heuristic written after an earlier failure, and the Solver retries with it in context~\citep{shinn2023reflexion}. The Extractor does not see which success conditions were missed, which discourages task-specific fixes in favor of generalizable guidance. The loop stops once the Solver succeeds $N_s$ consecutive times or fails $N_f$ times in total, with three possible outcomes: the heuristic is \emph{accepted} if the Solver failed at least once before these $N_s$ successes; the task is \emph{too easy} if the Solver never failed; and it is \emph{too hard} if the Solver reached $N_f$ failures. Whereas prior methods validate memories by the outcome of the trajectory they were drawn from~\citep{zhao2024expel,fu2024autoguide,allard2026experiential,ouyang2026reasoningbank} or by task feasibility~\citep{choi2026preping}, \daedalus{} thus tests the heuristic itself, and requiring $N_s$ consecutive successes rather than a single one lowers the probability of accepting an ineffective heuristic after a success due to chance (Appendix~\ref{sec:apdx-counterfactual}).

\begin{figure}[t]
    \centering
    \includegraphics[width=0.8\linewidth]{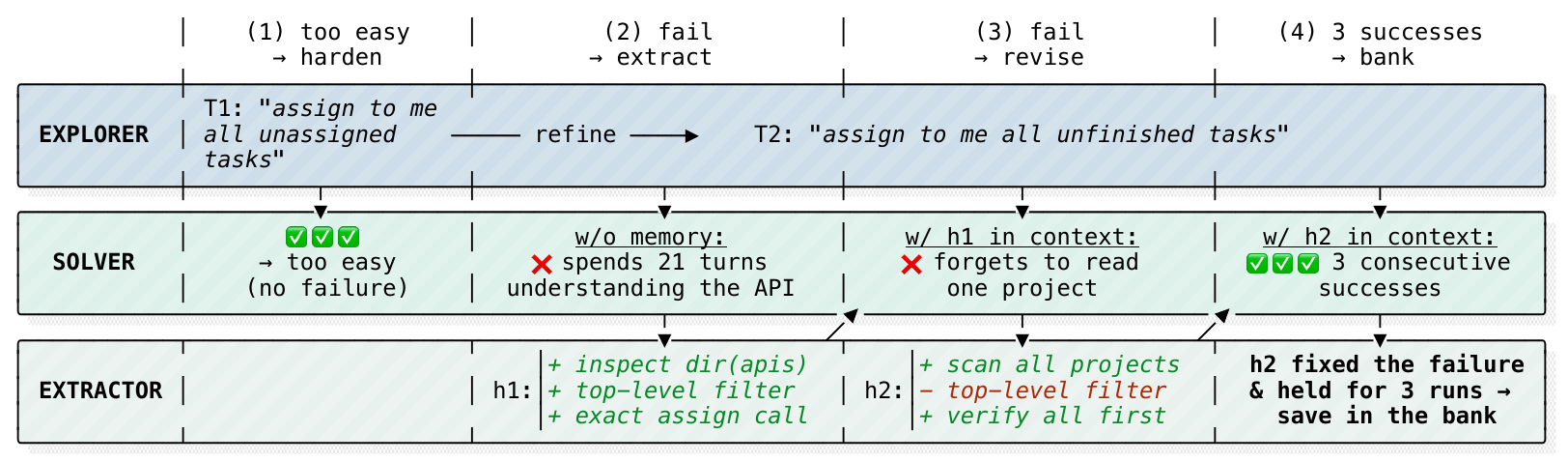}
    \caption{Timeline of AppWorld generation session~78. The Explorer first proposes a task that the Solver completes without failure, so the Explorer refines it to be harder. On the refined task, two failures yield successive heuristics $h_1$ and $h_2$. With $h_2$ in context, the Solver succeeds three consecutive times, validating its effectiveness; \(h_2\) is therefore accepted into the memory bank. Appendix~\ref{sec:example-session} shows the session in greater detail.}
    \label{fig:small-example-session}
\end{figure}

\paragraph{Calibrating tasks to the Solver.} In each session, the Explorer interacts with the environment, proposes a realistic task, and completes it itself to establish feasibility~\citep{shi2026taskcraft,ramrakhya2026scaling}. Because only accepted tasks yield heuristics, the Solver loop outcome doubles as a difficulty signal: when a task is too easy or too hard, the Explorer receives this outcome along with its previous attempt and revises the task to be harder or easier, up to $N_r$ times. A session that exhausts its refinements contributes no heuristic. As in proposer--solver curricula~\citep{yue2026dr}, this feedback steers generation toward the edge of the Solver's capability, where failures are recoverable. In our main experiments, the Solver shares the base agent's model, and we show in Section~\ref{sec:experiments} that the resulting bank also transfers to base agents from other model families. Figure~\ref{fig:small-example-session} shows how an example session unfolds.

\paragraph{Improving generation efficiency.} Two components reduce generation cost (Section~\ref{sec:ablation-generation}). Before the first session, a Surveyor explores the environment once and defines a target distribution of tasks across its areas, which the Explorer then follows to diversify the coverage of its tasks. In addition, after each session that required refinement, the refinement history is turned into \emph{guidelines} that help the Explorer calibrate difficulty in later sessions.

\paragraph{Consolidation and deployment.} After the last session, a Consolidator merges the accepted heuristics in a single LLM call, removing redundant or overlapping advice while preserving actionable guidance. The resulting bank is frozen and injected once into the base agent's context at the start of each test task; Section~\ref{sec:ablation-inference} compares this choice with per-turn retrieval.

\section{Experiments}\label{sec:experiments}

\subsection{Evaluation Framework}

\paragraph{Benchmarks.}

We run evaluations on three agentic benchmarks with different task types and environments: code-based app automation in AppWorld~\citep{trivedi2024appworld}, conversational customer service in $\tau^2$-bench~\citep{barres2026taubench}, and business workflows with SaaS applications in AutomationBench~\citep{shepard2026automationbench}. We evaluate agents on AppWorld's \textit{normal} test split, whose tasks use the same apps as the training split. On $\tau^2$-bench, we use only the \textit{retail} domain, since \textit{airline} contains ground-truth errors~\citep{rabanser2026towards}, and the dual-control setting of \textit{telecom} would require simulating user-side tools during exploration. Due to compute costs, we restrict AutomationBench to its \textit{Operations} split, whose higher public scores ensure a non-trivial baseline.

\paragraph{Baselines.}
We compare \daedalus{} against a no-memory baseline, which measures the capability of the base agent alone, and six memory methods. To our knowledge, PREPING~\citep{choi2026preping} is the only existing method that also builds memory from self-generated tasks instead of training tasks. The other five build memory from the benchmark training tasks, and therefore have access to information that \daedalus{} does not. ExpeL~\citep{zhao2024expel} and AutoGuide~\citep{fu2024autoguide} contrast successful and failed retries of each training task to extract global insights and context-specific guidelines, respectively. ERL~\citep{allard2026experiential} and ReasoningBank~\citep{ouyang2026reasoningbank} instead reflect on a single attempt per task, with the environment reward and an LLM judge as the respective feedback signals. ACE~\citep{zhang2026agentic} maintains an evolving playbook that it refines through incremental updates. Finally, to isolate the memory mechanism from task generation, we evaluate a \daedalus{}-curated variant that runs our Solver loop on the same training tasks.

\paragraph{Experimental setup.}

We split the public tasks of each benchmark into training tasks, from which we build reusable memory, and test tasks, on which we evaluate all methods. For AppWorld and $\tau^2$-bench retail, we use the official splits (90/168 and 74/40 training/test tasks). For AutomationBench Operations, we split the 100 tasks 30/70, stratified by the number of services each task involves. Memory generation never accesses test instances and only uses environment states available to the baselines using training tasks (Appendix~\ref{sec:apdx setup}).

To match the number of tasks available to the baselines, we run one \daedalus{} generation session per training task. The Explorer may refine each task up to $N_r=5$ times, and the Solver loop stops after $N_f=8$ failures. The loop accepts a heuristic after $N_s=3$ consecutive successes. We analyze these limits and the effect of accepted heuristics on their own tasks in Appendix~\ref{sec:appendix-generation-process}. We set these values from a few preliminary generation sessions on AppWorld and reuse them on $\tau^2$-bench and AutomationBench without tuning. The solver harness is a basic agent loop that executes Python code in AppWorld and uses standard function calls elsewhere, with no capabilities beyond what each environment natively provides.

All methods run a solver agent on tasks and delegate memory generation and retrieval to \textit{auxiliary} agents, such as the Surveyor, Explorer, Judge, Extractor, and Consolidator in \daedalus. At test time, we evaluate a \textit{base agent} that receives the resulting memory in context. We select two LLMs from the same family: the larger one for all auxiliary agents, and the smaller one for the solver and the base agent. By default, we use GPT-5.4 / GPT-5.4-mini on AppWorld and $\tau^2$-bench, and GPT-5.6 Terra / GPT-5.6 Luna on AutomationBench, which is a more challenging benchmark. Appendix~\ref{sec:apdx setup} lists the full configuration of each method.

For each method, we build one memory bank and average results over five independent inference runs, reporting the mean success rate (MSR) and \passpowerfive~\citep{yao2025taubench} with their standard errors. We also report the inference cost for one evaluation run at OpenAI standard-tier API prices, assuming perfect prompt caching.

\subsection{Main Results}

\begin{table}[t]
    \caption{Main results across five inference runs using a single memory bank per method and per benchmark. MSR denotes mean success rate, $\pm$ the standard error, and \$ the inference cost in USD per run, excluding memory generation. \textbf{Bold} marks the best result in each column, and \colorbox{Lavender}{shading} the best result among methods without training tasks.}
    \label{tab:main-results}
    \centering
    \begin{adjustbox}{width=\textwidth}
    \begin{tabular}{l c c c c c c c c c}
    \toprule

    & \multicolumn{3}{c}{AppWorld (normal)} & \multicolumn{3}{c}{$\tau^2$-bench (retail)} & \multicolumn{3}{c}{AutomationBench (Operations)}\\
    \cmidrule(lr){2-4}\cmidrule(lr){5-7}\cmidrule(lr){8-10}
    & MSR\up & \passpowerfive{}\up & \$\down
    & MSR\up & \passpowerfive{}\up & \$\down
    & MSR\up & \passpowerfive{}\up & \$\down \\

    \midrule

        \emph{No-memory baseline}
                  & 44.3\conf{1.1} & 14.9\confpass{2.8} & 3.2
                  & 57.5\conf{1.6} & 22.5\confpass{6.6} & \textbf{0.5}
                  & 31.7\conf{3.2} & 12.9\confpass{4.0} & 1.0 \\

    \midrule

    \rowcolor{gray!15}
    \multicolumn{10}{c}{\textbf{\textit{Methods using training tasks}}} \\
    AutoGuide     & 44.0\conf{0.5} & 17.9\confpass{3.0} & 7.6
                  & 61.5\conf{2.3} & 27.5\confpass{7.1} & 3.2
                  & 35.4\conf{1.7} & 18.6\confpass{4.6} & 4.7 \\
    ReasoningBank & 48.0\conf{1.0} & 19.6\confpass{3.1} & 3.3
                  & 64.0\conf{2.9} & 32.5\confpass{7.5} & \textbf{0.5}
                  & 19.1\conf{1.7} & 2.9\confpass{2.0} & \textbf{0.7} \\
    ERL           & 58.6\conf{1.3} & 29.2\confpass{3.5} & 23.0
              & 64.0\conf{3.6} & 30.0\confpass{7.2} & 5.1
                  & 33.4\conf{1.5} & 15.7\confpass{4.3} & 3.2 \\
    ExpeL         & 59.0\conf{1.9} & 33.3\confpass{3.6} & 4.6
                  & 67.0\conf{1.8} & \textbf{37.5}\confpass{7.8} & 0.9
                  & \textbf{42.6}\conf{0.5} & 21.4\confpass{4.9} & 1.5 \\
    ACE           & 60.5\conf{0.8} & \textbf{41.1}\confpass{3.8} & 7.6
                  & \textbf{70.5}\conf{3.8} & \textbf{37.5}\confpass{7.8} & 2.2
                  & 26.3\conf{1.9} & 10.0\confpass{3.6} & 0.8 \\
    \textbf{\daedalus{}-curated}
                  & \textbf{60.8}\conf{0.9} & 36.3\confpass{3.7} & \textbf{3.0}
                  & 70.0\conf{3.6} & 35.0\confpass{7.5} & 0.6
                  & 40.0\conf{2.3} & \textbf{24.3}\confpass{5.1} & 1.0 \\

    \midrule

    \rowcolor{gray!15}
    \multicolumn{10}{c}{\textbf{\textit{Methods without training tasks}}} \\

    PREPING       & 56.0\conf{1.4} & 25.6\confpass{3.4} & 4.0
                  & 60.0\conf{1.8} & 25.0\confpass{6.9} & 0.7 & 34.9\conf{1.5} & 15.7\confpass{4.3} & \colorbox{Lavender}{1.1} \\
    \textbf{\daedalus{}}
                  & \colorbox{Lavender}{60.2\conf{0.9}} & \colorbox{Lavender}{32.1\confpass{3.6}} & \colorbox{Lavender}{3.2}
                  & \colorbox{Lavender}{67.5\conf{2.5}} & \colorbox{Lavender}{\textbf{37.5}\confpass{7.8}} & \colorbox{Lavender}{\textbf{0.5}}
                  & \colorbox{Lavender}{36.0\conf{1.5}} & \colorbox{Lavender}{21.4\confpass{4.9}} & 1.2 \\
                  \bottomrule
        \end{tabular}
    \end{adjustbox}
\end{table}

\paragraph{\daedalus{} is the strongest method without training tasks.}
\daedalus{} raises the MSR of the base agent by 15.9, 10.0 and 4.3 points over the no-memory baseline on AppWorld, $\tau^2$-bench and AutomationBench (Table~\ref{tab:main-results}). \passpowerfive increases by 1.7$\times$ to 2.2$\times$, indicating that the method also improves consistency across runs. \daedalus{} also consistently outperforms PREPING on both metrics. On AppWorld and $\tau^2$-bench, it does so at no extra inference cost over the no-memory baseline, mainly because the agent needs fewer turns (Figure~\ref{fig:transferability}).

\paragraph{\daedalus{} is competitive with methods that use curated tasks.}
Without any training task, \daedalus{} is within error margins of the best method using them in four of the six MSR and \passpowerfive{} columns, falling behind only ACE on AppWorld \passpowerfive{} and ExpeL on AutomationBench MSR. It is also robust across benchmarks as it never falls below the no-memory baseline, unlike ACE, ReasoningBank and AutoGuide. Given the same training tasks, \daedalus{}-curated ranks first within error margins on every benchmark and metric, at one of the lowest inference costs, which shows that our memory mechanism is competitive independently of task generation. Self-generated tasks recover most of the benefit of curated ones, since \daedalus underperforms its curated variant by only 0.6, 2.5 and 4.0 MSR points, and the \passpowerfive{} gap is within error margins on all three.

\paragraph{Heuristics generated by one model family transfer to another.}
We investigate to what extent \daedalus{} generalizes to various model families, and whether a bank generated with one model family also helps base agents from another. We therefore run \daedalus{} on AppWorld with auxiliary / solver model pairs from the Qwen and DeepSeek model families, and evaluate each memory bank with each base agent. Table~\ref{tab:transfer} shows that all nine gains are positive, indicating that a base agent does not need the bank of its own model family. For example, GPT-5.4-mini gains as much from the Qwen bank as from its own (+16.8 against +15.9). DeepSeek-V4-Flash gains less (3.0 to 8.3 points), as its no-memory baseline of 80.6\% leaves less room for improvement. Figure~\ref{fig:transferability} extends this result to more base agent LLMs. With \daedalus{} memory, they all reach a higher MSR in fewer turns, showing that our method benefits a wide range of models.
\begin{table}[h]
  \centering
  \small
  \setlength{\tabcolsep}{5pt}
  \caption{Cross-family transfer of \daedalus heuristics on AppWorld. First row: MSR of each base agent (column) without memory. Others: MSR improvement when the base agent uses the \daedalus heuristic bank generated by an auxiliary / solver pair (row).}
  \label{tab:transfer}
  \begin{tabular}{ll ccc}
    \toprule
    \multicolumn{2}{l}{\textit{\daedalus memory}} & \multicolumn{3}{c}{\textit{Base agent}} \\
    \cmidrule(r){1-2}\cmidrule(l){3-5}
    Auxiliary & Solver & GPT-5.4-mini & Qwen3.6-35B-A3B & DeepSeek-V4-Flash \\
    \midrule
    \multicolumn{2}{l}{\emph{No-memory baseline}} & 44.3\conf{1.1} & 44.8\conf{1.3} & 80.6\conf{1.6} \\
    \midrule
    GPT-5.4          & GPT-5.4-mini      & +15.9\conf{1.4} & +16.1\conf{1.6} & +8.3\conf{1.8} \\
    Qwen3.8-Flash    & Qwen3.6-35B-A3B   & +16.8\conf{2.4} & +29.3\conf{1.7} & +4.9\conf{2.0} \\
    DeepSeek-V4-Pro  & DeepSeek-V4-Flash & +11.4\conf{3.6} & +14.5\conf{1.3} & +3.0\conf{3.0} \\
    \bottomrule
  \end{tabular}
\end{table}

\section{Analysis and Ablations}\label{sec:pipeline-ablation}

This section examines each stage of \daedalus{}. For memory generation (Section~\ref{sec:ablation-generation}), we ablate the pipeline components, validate the LLM judge against official benchmark verifiers, and vary the exploration budget and the auxiliary models. For inference (Section~\ref{sec:ablation-inference}), we compare consolidating the bank and injecting it whole with retrieving heuristics at each turn. Finally, we test whether the generated tasks can serve as a proxy test set for ranking models (Section~\ref{sec:evaluation-proxies}). Unless noted otherwise, all analyses use AppWorld, with GPT-5.4 for the auxiliary agents and GPT-5.4-mini for the Solver and base agent.

\subsection{Memory Generation}\label{sec:ablation-generation}

\paragraph{Pipeline ablation.}

We perform a cumulative ablation on AppWorld to assess how each component of \daedalus affects downstream performance and generation efficiency (Table~\ref{tab:pipeline-ablation}). Starting with heuristics extracted from a single 100-turn Explorer trajectory \textsc{(a)}, we switch to multiple 40-turn sessions with access to earlier heuristics \textsc{(b)}. In \textsc{(c)}, the Explorer proposes tasks for a Solver to attempt, and heuristics are extracted directly from all Solver trajectories, without retry-based validation. We add failure-triggered revisions and repeated-success validation in \textsc{(d)}, guidelines derived from task refinements in \textsc{(e)}, and a preliminary environment survey in \textsc{(f)}, yielding the complete \daedalus{} pipeline. All multi-session variants use 90 sessions. Two findings emerge:

\begin{table}[t]
  \centering
  \caption{Cumulative ablation of the \daedalus{} generation pipeline on AppWorld. Each row adds its component to the preceding configuration; Accept. \% is the percentage of 90 sessions ending with an accepted heuristic; \# Refin. counts task revisions across sessions; Inf. \$ and Gen. \$ denote the inference and one-time memory generation costs, respectively.}
  \label{tab:pipeline-ablation}
  \begin{adjustbox}{width=\textwidth}
    \begin{tabular}{l c c c c c c}
    \toprule
         & MSR\up
         & \passpowerfive{}\up
         & Inf. \$\down
         & Accept. \%\up
         & \# Refin.\down
         & Gen. \$\down \\
    \midrule
    No-memory baseline
         & 44.3\conf{1.1}
         & 14.9\confpass{2.8}
         & 3.2
          & --
          & --
         & -- \\
    \midrule
    Single Explorer \textsc{(a)}
         & 35.7\conf{0.8}
         & 10.1\confpass{2.3}
         & 3.5
          & --
          & --
         & 5.2 \\

    + Multi-session Explorer \textsc{(b)}
     & 38.7\conf{3.6}
     & 11.3\confpass{2.4}
     & 2.9
      & --
      & --
     & 92.0 \\

    + Explorer tasks and Solver traces \textsc{(c)}
     & 54.5\conf{2.8}
     & 25.0\confpass{3.3}
     & \textbf{1.9}
      & --
      & --
     & 46.6 \\
     
    + Solver loop \textsc{(d)}
         & 59.0\conf{1.4}
         & 30.4\confpass{3.6}
         & 3.3
          & 86.7
          & 183
         & 210.4 \\

    + Explorer guideline bank \textsc{(e)}
         & 57.9\conf{2.1}
         & 31.5\confpass{3.6}
         & 3.2
          & 91.1
          & 161
         & 185.3 \\

    + Environment survey (\daedalus) \textsc{(f)}
         & \textbf{60.2}\conf{0.9}
         & \textbf{32.1}\confpass{3.6}
         & 3.2
          & \textbf{97.8}
          & \textbf{118}
         & 109.7 \\
    \bottomrule
    \end{tabular}
    \end{adjustbox}
\end{table}

\textbf{\textsc{(i)} Solver-grounded heuristics drive downstream performance.} Exploration alone is insufficient: variant \textsc{(a)} underperforms the no-memory baseline, and the additional cost of multiple sessions does not allow \textsc{(b)} to overcome that loss. Two substantial jumps appear along the success axis. The first one (+15.8 MSR) comes from including Solver traces \textsc{(c)}, showing that the qualitative content to derive heuristics lies in attempt trajectories. The second jump (+4.5 MSR) lands when validating heuristics with our Solver loop, which empirically removes noisy lessons from the bank, a result that is consistent with the scores of \daedalus{}-curated in Table~\ref{tab:main-results}.

\begin{wrapfigure}{r}{0.45\textwidth}
  \centering
  \vspace{-1\baselineskip}
  \includegraphics[width=\linewidth]{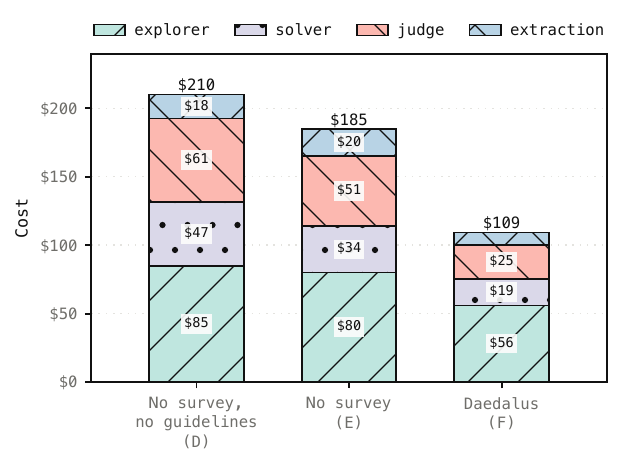}
  \caption{Generation cost by agent role for \textsc{(d)}, \textsc{(e)}, and the complete pipeline.}
  \label{fig:cost-distrib-generation}
  \vspace{-2\baselineskip}
\end{wrapfigure}
\textbf{\textsc{(ii)} Task design guidelines and the environment survey improve generation efficiency.} Guidelines in \textsc{(e)} carry refinement lessons into later sessions, helping the Explorer reach the target difficulty. They improve cost-efficiency rather than memory quality, reducing refinements from 183 to 161 and generation cost by 12\%, while MSR changes by $-1.1$ points, within standard error (Table~\ref{tab:pipeline-ablation}). The survey in \textsc{(f)} gives the Explorer a target task distribution over environment areas, reducing the exploration needed to select each task. Together, these components increase the share of sessions yielding an accepted heuristic and cut generation cost by 48\% relative to \textsc{(d)}. Figure~\ref{fig:cost-distrib-generation} shows that Solver and Judge costs fall by 60\% due to fewer refinements.

\paragraph{LLM judge validation.} \label{sec:judge-validation}

A mistaken judgment can misdirect task refinement or validate an unreliable heuristic. To assess our LLM judge, we compare it against benchmark-provided verifiers on official test sets.
For each task, we let our Explorer inspect the environment and restate the benchmark's oracle requirements as natural-language success conditions, as it would for a task it generated. The judge then receives the task instruction, these conditions, and the Solver trajectory, but not the official verifier or its verdict. For each benchmark with $N$ tasks, we judge one Solver trajectory per task five times and measure agreement with the official verifier ($\kappa_{\mathrm{bench}}$) and between repeated judgments ($\kappa_{\mathrm{inter}}$). Table~\ref{tab:judge-validation} shows substantial agreement with the official verifiers ($\kappa_{\mathrm{bench}} > 0.7$) and consistent verdicts across calls ($\kappa_{\mathrm{inter}} > 0.8$). Precision exceeds recall on every benchmark, so the judge leans toward rejecting successes rather than accepting failures, which is the safer direction for accepting heuristics.

\begin{table}[t]
\centering
\caption{Agreement between the LLM judge and official verifiers, and across five judge calls per trajectory. Precision and recall treat success as positive.}
\label{tab:judge-validation}

\small
\setlength{\tabcolsep}{4pt} 
\begin{tabular}{@{}lccccc@{}}
\toprule
Benchmark & $N$ & Precision & Recall
          & $\kappa_{\mathrm{bench}}$
          & $\kappa_{\mathrm{inter}}$ \\
\midrule
AppWorld        & 168 & .943 & .846 & .807 & .848 \\
$\tau^2$-bench  & 40  & .889 & .842 & .749 & 1.000 \\
AutomationBench & 70  & .863 & .807 & .729 & .902 \\
\bottomrule
\end{tabular}
\end{table}

\paragraph{Scaling with the exploration budget.}

\begin{wrapfigure}{r}{0.45\textwidth}
  \vspace{-1em}
  \centering
  \captionsetup{font=small, skip=4pt}
  \includegraphics[width=\linewidth]{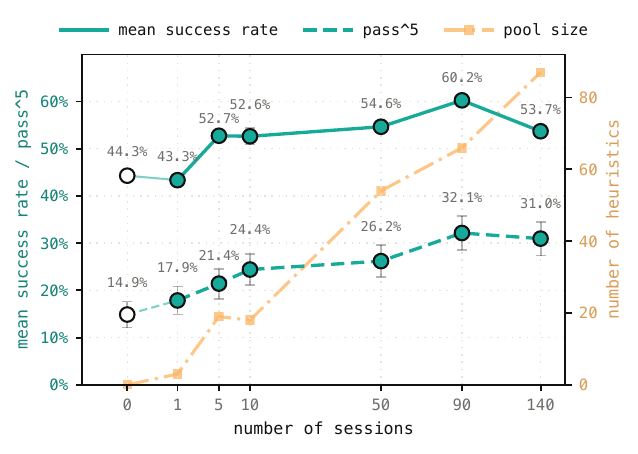}
  \caption{Scaling with the exploration budget on AppWorld, over a single generation run. Error bars are standard errors over five inference runs.}
  \label{fig:num-sessions}
\end{wrapfigure}

We study how performance depends on the exploration budget by continuing a generation run up to 140 sessions and evaluating the memory at successive checkpoints (Figure~\ref{fig:num-sessions}). Five sessions already deliver more than half of the improvement reached at the 90-session peak. Beyond 90 sessions, MSR decreases even as the bank grows. A hierarchical consolidation, designed to scale to large banks, does not recover the loss (54.6 vs.\ 53.7 MSR): with a larger bank, consolidation misstates the scope of some heuristics, and such minor changes can have a large impact on specific parts of the test distribution (Appendix~\ref{app:hierarchical-consolidation}). The early gains matter more in practice, since they make a useful memory affordable after only a few sessions.

\begin{wraptable}[8]{r}{0.45\textwidth}
    \vspace{-1em}
    \footnotesize
    \setlength{\tabcolsep}{3.5pt}
    \renewcommand{\arraystretch}{1.0}
    \caption{Effect of auxiliary model choice.}
    \begin{tabular}{l c c c}
        \toprule
        Aux.\ model & MSR\up & \passpowerfive{}\up & Gen. \$\down\\
        \midrule
        \emph{no memory} & 44.3\conf{1.1} & 14.9\confpass{2.8} & 0\\
        \midrule
        GPT-5.4-mini & 51.2\conf{1.6} & 20.2\confpass{3.1} & 67.2 \\
        GPT-5.4      & \textbf{60.2}\conf{0.9} & \textbf{32.1}\confpass{3.6} & 109.7\\
        \bottomrule
     \end{tabular}
     \label{tab:aux-model}
\end{wraptable}
\paragraph{Effect of model selection.}
In our default setup, auxiliary agents run on a larger model than the Solver and base agent. To test whether \daedalus{} requires this asymmetry, we run every auxiliary agent with GPT-5.4-mini (Table~\ref{tab:aux-model}). This setup still outperforms the no-memory baseline by 6.9 MSR and 5.3 \passpowerfive points, about half of the gain obtained with GPT-5.4. A stronger generation pipeline thus yields better memory, but \daedalus{} remains useful when a larger model is not available.

\subsection{Memory Use at Inference\label{sec:ablation-inference}}

\begin{figure}[t]
\centering
\begin{minipage}[t]{0.48\textwidth}
    \centering
    \footnotesize
    \setlength{\tabcolsep}{3pt}
    \captionof{table}{Consolidation and injection of the heuristic bank on AppWorld. \textbf{Bold}: best; \underline{underline}: second best.}
    \label{tab:memory-use}
    \begin{adjustbox}{width=\textwidth}
        \begin{tabular}{l c c c c r}
            \toprule
            Policy & Dedup. & Consol. & MSR\up & \passpowerfive{}\up & \$\down \\
            \midrule
            No memory & -- & -- & 44.3\conf{1.1} & 14.9\confpass{2.8} & 3.2 \\
            \midrule
            \rowcolor{gray!15}
            \multicolumn{6}{l}{\textit{Top-5 per turn}} \\
            Random     & \cmark & \cmark & 51.7\conf{1.5} & 19.0\confpass{3.0} & 4.1 \\
            BM25       & \cmark & \cmark & 52.0\conf{1.3} & 22.0\confpass{3.2} & 3.8 \\
            Qwen3-Emb. & \cmark & \cmark & \underline{54.3}\conf{1.6} & 24.4\confpass{3.3} & 3.7 \\
            \midrule
            \rowcolor{gray!15}
            \multicolumn{6}{l}{\textit{Whole bank per turn}} \\
            All        & \cmark & \cmark & 51.5\conf{1.2} & \underline{28.0}\confpass{3.5} & 3.5 \\
            \midrule
            \rowcolor{gray!15}
            \multicolumn{6}{l}{\textit{Whole bank at start}} \\
            All        & \xmark & \xmark & 49.5\conf{1.0} & 26.2\confpass{3.4} & 4.9 \\
            All        & \cmark & \xmark & 49.8\conf{2.5} & 27.4\confpass{3.5} & 4.8 \\
            \daedalus{} & \cmark & \cmark & \textbf{60.2}\conf{0.9} & \textbf{32.1}\confpass{3.6} & \textbf{3.2} \\
            \bottomrule
        \end{tabular}
    \end{adjustbox}
\end{minipage}%
\hfill%
\begin{minipage}[t]{0.48\textwidth}
    \centering
    \vspace{2pt}
    \includegraphics[width=\linewidth]{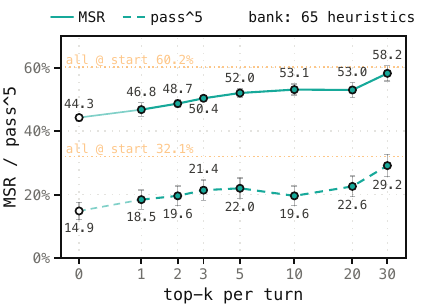}
    \captionof{figure}{BM25 retrieval per turn with increasing $k$. Dotted lines: whole bank at start; $k{=}0$: no memory.} 
    \label{fig:topk-success}
\end{minipage}
\end{figure}

Heuristics accepted in separate sessions often overlap, which raises two design questions at inference: whether to consolidate them into a compact bank, and whether to inject that whole bank or retrieve some heuristics at each turn. Injected at task start, the consolidated bank is both the best-performing and the cheapest option, improving MSR by 10.7 points over the unconsolidated heuristics while considerably shortening the context; deduplication alone yields no significant gain (Table~\ref{tab:memory-use}). We then test whether retrieval can select more relevant heuristics per turn, comparing BM25, Qwen3-Embedding-4B~\citep{zhang2025qwen3}, and a random control, each retrieving five heuristics per turn (Appendix~\ref{app:retrieval-details}). All three improve over the no-memory baseline by up to 10 MSR points but do not differ significantly. A plausible reason is that useful heuristics are not necessarily semantically similar to the current step: retrieving them is an oblique retrieval problem, in which relevance is latent rather than expressed in the text~\citep{tchuindjo2026obliq}. In line with this, retrieving more heuristics with BM25 improves success, but stays below whole-bank injection (58.2\% at $k{=}30$; Figure~\ref{fig:topk-success}). At this scale, injecting the whole bank once at task start is thus the most effective option we tested, while larger banks would call for retrievers suited to oblique queries.

\begin{wrapfigure}[15]{r}{0.45\textwidth}
  \vspace{-1\baselineskip}
  \centering
  \includegraphics[width=\linewidth]{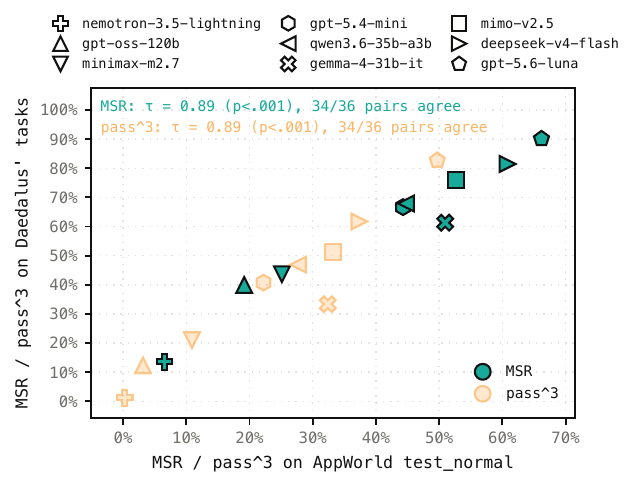}
  \caption{Performance on \daedalus-generated tasks versus the AppWorld test set, for nine models evaluated without memory.}
  \label{fig:test-set-correlation}
  \vspace{-1\baselineskip}
\end{wrapfigure}

\subsection{Self-Generated Tasks as Evaluation Proxies}\label{sec:evaluation-proxies}
Environments without training tasks usually lack a test set as well, which makes it hard to compare agents on them. We test whether the tasks generated by \daedalus{} can fill this role by evaluating nine models without memory on the 81 accepted tasks of the 90-session AppWorld run and on the AppWorld \textit{normal} test split, with three runs per task set. Rankings are strongly aligned for both MSR and pass\textasciicircum{}3 (Kendall's $\tau=0.89$, $p<.001$ for both), with 34 of 36 model pairs ordered consistently in each case (Figure~\ref{fig:test-set-correlation}). Absolute success rates differ, as the generated tasks were calibrated to GPT-5.4-mini, but the relative order of models is preserved. These results suggest that \daedalus{}-generated tasks can serve as a proxy for ranking models when no curated test set is available.

\section{Conclusion} 

Memory-based agents usually learn from curated tasks and an existing verifier, which a new environment does not provide. \daedalus{} removes the need for both: the agent builds its memory by practicing on calibrated tasks it generates itself, and keeps a heuristic only once it has turned one of the solver's failures into repeated success. Across three benchmarks, the resulting memory is the strongest among methods without training tasks and competitive with methods that require them, at marginal or no additional inference cost. The main reason behind this improvement is validation: heuristics drawn from exploration alone hurt the agent, whereas those grounded in its failures and confirmed empirically improve it. Additionally, \daedalus{} generates memory that transfers across model families, and tasks that rank models consistently with curated benchmarks, suggesting that it can also supply evaluation data where none exists.

On AppWorld, performance declined beyond 90 sessions as the memory grew, which raises open questions for scaling: how the composition and coverage of the bank vary across environments, how heuristics should be consolidated as the bank grows, and which retrieval mechanisms can best exploit large memories at inference time.
\daedalus{} also builds its bank entirely before deployment and relies on an imperfect LLM judge; updating memory at test time and developing more reliable multi-judge validation schemes are natural next steps.
We release our code and all generated artifacts to encourage further work along these directions.

\newpage

\FloatBarrier

\newpage

\subsection*{Acknowledgements}

We thank Illuin Technology for providing the time and resources needed to carry out this research. This work was conducted within LIAGORA, a joint laboratory (``LabCom'') between Illuin Technology and the MICS laboratory of CentraleSupélec, supported by the French National Research Agency (ANR). We are also grateful to Quentin Macé for his thoughtful reading of the manuscript and constructive feedback.

\subsection*{Author Contributions}

Antoine Edy, Max Conti, and Victor Xing framed the project, designed the method, and implemented and ran the generation experiments. Antoine implemented and ran benchmark inference experiments. He carried out the memory-generation pipeline ablations, studied the effect of the exploration budget, compared model rankings on generated and official AppWorld tasks with Victor, and co-wrote the manuscript. Max shaped the methodology, studied memory-injection methods at inference, and co-wrote the paper. Victor implemented and ran the transferability experiments and benchmark inference on $\tau^2$-bench and AutomationBench. He also evaluated smaller auxiliary models, ran counterfactual replays of accepted heuristics, co-wrote the manuscript, and supervised the project. Marc-Antoine Allard implemented and ran benchmark inference experiments and the test-time retrieval ablations. He contributed to research ideation in the later stages of the project and wrote parts of the paper, including the related work. Nawfal Benhamdane contributed to research ideation, implemented and ran AppWorld experiments, carried out the pipeline ablations with Antoine, conducted the LLM-judge study, and reviewed the manuscript. Gautier Viaud supervised the project and contributed to its administration and funding.

\subsection*{AI use statement}

We used generative AI tools to aid and polish the writing, draft sections of the paper, retrieve and discover related work, and support research ideation and execution, including code generation. We also used LLM agents to generate synthetic datasets: the tasks produced by \daedalus are the output of the method under study, serve as the source of its memory, and are used as an evaluation set in Section~\ref{sec:pipeline-ablation}. Their generation is described in Section~\ref{sec:method}. We did not use generative AI tools to prove mathematical claims, as this work contains none. The authors carefully reviewed all AI-assisted work. Generated code was reviewed by the authors, including benchmark and baseline reimplementations, and each cited work was manually checked to confirm its existence. We take full responsibility for the final content of this work, including text, claims, and artifacts produced with the aid of generative AI.

\subsection*{Reproducibility statement}

The generation pipeline is described in Section~\ref{sec:method} and formalized in Appendix~\ref{apdx pseudo code}, and Appendix~\ref{sec:apdx setup} gives the configuration of every method, including the models, splits and hyperparameters. We release the code, the prompts, all the generated tasks, heuristic banks and agent trajectories, so that every experiment can be rerun or inspected: \href{https://github.com/illuin-tech/daedalus}{\faicon{github} \texttt{\small www.github.com/illuin-tech/daedalus}}.

\newpage

\bibliography{bibfile}
\bibliographystyle{bibstyle}

\newpage

\appendix
\raggedbottom

\section{Pseudo-code algorithms} \label{apdx pseudo code}

We formalize the \daedalus memory generation phase (Algorithms~\ref{alg:main}--\ref{alg:solver}), its supervised counterpart \daedalus-curated (Algorithm~\ref{alg:accumulation}), and memory-augmented inference (Algorithm~\ref{alg:inference}).

$\mathcal{E}$ is a collection of resettable environments, and $e \gets \envop{Reset}(\mathcal{E})$ selects an environment and restores its initial state. A task is $\tau=(u,\sigma)$, where $u$ is its instruction and $\sigma$ its success conditions. The agents are $\agentcall{Surveyor}$, $\agentcall{Explorer}$, $\agentcall{Solver}$, $\agentcall{Judge}$, $\agentcall{Extractor}$, and $\agentcall{Consolidator}$. The $\agentcall{Surveyor}$ returns a survey report $\mathcal{S}$ and coverage state $\mathcal{C}$. $\mathcal{T}$ contains accepted tasks, $\mathcal{H}$ accepted heuristics, and $\mathcal{G}$ task design guidelines. $\varnothing$ denotes absence and $[\,]$ an empty ordered history. The generation hyperparameters are $n$ sessions, $N_s$ consecutive successes, $N_f$ cumulative failures, and $N_r$ task refinements.

Sans-serif names denote prompted LLM calls (possibly multi-turn interactions with an environment), small-cap names algorithms defined below, and monospace names environment operations.

\subsection{\daedalus generation}

\begin{algorithm}[H]
\caption{\daedalus{} memory generation}
\label{alg:main}
\begin{algorithmic}[1]
\Require environments $\mathcal{E}$; hyperparameters $n,N_s,N_f,N_r$
\Ensure frozen heuristic bank $\mathcal{H}^{\star}$
\State $(\mathcal{S}, \mathcal{C}) \gets \agentcall{Surveyor}(\mathcal{E})$
\State $\mathcal{T} \gets [\,]$;\; $\mathcal{H} \gets [\,]$;\;
       $\mathcal{G} \gets [\,]$
\For{$i = 1, \dots, n$}
  \State $(\tau,h,o,\mathcal{L}) \gets \textsc{GenerateHeuristic}(\mathcal{E},
         \mathcal{S},\mathcal{T},\mathcal{G},\mathcal{C},N_s,N_f,N_r)$
    \Comment{Algorithm~\ref{alg:session}}
  \If{$o = \texttt{accepted}$}
    \State append $\tau$ to $\mathcal{T}$;\; append $h$ to $\mathcal{H}$;\;
           update $\mathcal{C}$ with the tags of $\tau$
  \EndIf
  \If{$|\mathcal{L}| > 1$}
    \State $\mathcal{G} \gets \agentcall{CreateGuidelines}
           (\mathcal{G},\mathcal{L})$
  \EndIf
\EndFor
\State \Return $\mathcal{H}^{\star} \gets \agentcall{Consolidator}(\mathcal{H})$
\end{algorithmic}
\end{algorithm}

\begin{algorithm}[H]
\caption{\textsc{GenerateHeuristic}: propose a task, calibrate it, and mine a heuristic}
\label{alg:session}
\begin{algorithmic}[1]
\Require environments $\mathcal{E}$; survey $\mathcal{S}$; accepted tasks $\mathcal{T}$;
         guidelines $\mathcal{G}$; coverage $\mathcal{C}$; hyperparameters $N_s,N_f,N_r$
\Ensure $(\tau,h,o,\mathcal{L})$: task, candidate heuristic, outcome, and session history
\State $(\tau,\xi) \gets \agentcall{Explorer.Explore}(
       \mathcal{E},\mathcal{S},\mathcal{T},\mathcal{G},\mathcal{C})$
  \Comment{$\xi$ is the explorer trajectory}
\State $\mathcal{L} \gets [\,]$
\State $(h,o) \gets \textsc{SolverLoop}(\mathcal{E},\tau,N_s,N_f)$
  \Comment{Algorithm~\ref{alg:solver}}
\State append $(\tau,\xi,o)$ to $\mathcal{L}$
\State \textbf{if} $o = \texttt{accepted}$ \textbf{then return}
       $(\tau,h,o,\mathcal{L})$
\For{$j = 1, \dots, N_r$}
  \State $d \gets \texttt{harder}$ \textbf{if} $o = \texttt{too\_easy}$
         \textbf{else} \texttt{easier}
  \State $(\tau,\xi) \gets \agentcall{Explorer.Refine}(
         \mathcal{E},\tau,\xi,d,\mathcal{L},\mathcal{G},\mathcal{C})$
  \State $(h,o) \gets \textsc{SolverLoop}(\mathcal{E},\tau,N_s,N_f)$
    \Comment{Algorithm~\ref{alg:solver}}
  \State append $(\tau,\xi,o)$ to $\mathcal{L}$
  \State \textbf{if} $o = \texttt{accepted}$ \textbf{then return}
         $(\tau,h,o,\mathcal{L})$
\EndFor
\State \Return $(\tau,h,o,\mathcal{L})$
\end{algorithmic}
\end{algorithm}

\begin{algorithm}[H]
\caption{\textsc{SolverLoop}: mine and validate a heuristic}
\label{alg:solver}
\begin{algorithmic}[1]
\Require environments $\mathcal{E}$; task $\tau=(u,\sigma)$; hyperparameters $N_s,N_f$
\Ensure $(h,o)$: heuristic (if accepted) and outcome
\State $h \gets \varnothing$;\; $\mathit{streak} \gets 0$;\;
       $\mathit{failures} \gets 0$
\Repeat
  \State $e \gets \envop{Reset}(\mathcal{E})$;\;
         $\zeta \gets \agentcall{Solver}(e,u,h)$
  \State $\mathit{passed} \gets \agentcall{Judge}(\zeta,u,\sigma)$
  \If{$\mathit{passed}$}
    \State $\mathit{streak} \gets \mathit{streak}+1$
  \Else
    \State $\mathit{streak} \gets 0$;\;
           $\mathit{failures} \gets \mathit{failures}+1$
    \State $h \gets \agentcall{Extractor.Edit}(u,\zeta,h)$
      \Comment{edits or extends $h$; overall failure only}
  \EndIf
\Until{$\mathit{streak} = N_s$ \textbf{or} $\mathit{failures} = N_f$}
\State \textbf{if} $\mathit{streak} = N_s$ \textbf{and} $h = \varnothing$
       \textbf{then return} $(\varnothing,\texttt{too\_easy})$
\State \textbf{if} $\mathit{streak} = N_s$ \textbf{then return}
       $(h,\texttt{accepted})$
\State \Return $(h,\texttt{too\_hard})$
\end{algorithmic}
\end{algorithm}

\subsection{\daedalus-curated}

\begin{algorithm}[H]
\caption{Accumulation from labeled tasks (\daedalus-curated)}
\label{alg:accumulation}
\begin{algorithmic}[1]
\Require environments $\mathcal{E}$; labeled tasks $\mathcal{Q}$ with verifier $V$;
         hyperparameters $N_s,N_f$
\Ensure frozen heuristic bank $\mathcal{H}^{\star}$
\State $\mathcal{H} \gets [\,]$
\For{$\tau \in \mathcal{Q}$}
  \State $(h,o) \gets \textsc{SolverLoop}(\mathcal{E},\tau,N_s,N_f)$
    \Comment{Algorithm~\ref{alg:solver}; $\agentcall{Judge}$ replaced by $V$}
  \State \textbf{if} $o = \texttt{accepted}$ \textbf{then append} $h$
         \textbf{to} $\mathcal{H}$
\EndFor
\State \Return $\mathcal{H}^{\star} \gets \agentcall{Consolidator}(\mathcal{H})$
\end{algorithmic}
\end{algorithm}

\subsection{Memory-augmented inference}

\begin{algorithm}[H]
\caption{Memory-augmented inference (\textsc{All @ Start})}
\label{alg:inference}
\begin{algorithmic}[1]
\Require environments $\mathcal{E}$; frozen bank $\mathcal{H}^{\star}$;
         instruction $u$
\Ensure Solver trajectory $\zeta$
\State $e \gets \envop{Reset}(\mathcal{E})$
\State $\zeta \gets \agentcall{Solver}(e,u,\mathcal{H}^{\star})$
\State \Return $\zeta$
\end{algorithmic}
\end{algorithm}

\section{Limitations} \label{app:limitations}

\daedalus{} builds a memory bank once, before deployment, from a single environment, and relies on an LLM judge checking self-written success conditions as its only success signal. This design makes memory creation fully autonomous but leaves some limitations open.

\paragraph{Memory lifecycle.} We study memory construction as a one-time stage and keep the bank frozen at test time. Continuing to update it from deployment experience, for instance by alternating \daedalus{} sessions with test-time reflection, is left to future work; our preliminary attempts at such iterative schemes did not yield consistent gains.

\paragraph{Environment and verification coverage.} \daedalus{} assumes that the explorer can set up and complete tasks alone in a resettable environment. Dual-control settings such as $\tau^2$-bench telecom, where a simulated user must also act on the environment, are not yet supported. Success is also judged by an LLM against self-written conditions, whose agreement with official verifiers is substantial but imperfect ($\kappa_{\text{bench}}$ of 0.73--0.81). For code-based tasks, an executable verifier agent could replace or complement the judge.

\paragraph{Consolidation at scale.} \label{app:hierarchical-consolidation}
To test whether the 140-session bank is too large for a single consolidation call, we also consolidate its 130 heuristics hierarchically: we split them at random into eight groups and merge these pairwise with the same prompt (Prompt~\ref{prompt:consolidation}) until one bank remains ($8\to4\to2\to1$). Both 140-session banks fall below the 90-session bank overall, yet on the 131 \emph{Other} tasks they nearly match it (Table~\ref{tab:hierarchical-consolidation}, shaded): most of the gap sits in three small categories, where a few heuristics lost their scope. The single call narrows the 90-session rule to \emph{``enumerate every matching container''}, so the Solver reads a single page of playlists or friends; the hierarchical merge broadens a note on public playlists into locating any playlist with \texttt{is\_public=True}, which hides private ones.

\begin{table}[H]
  \centering
  \caption{Consolidating the 140-session bank on AppWorld by task category (number of tasks in parentheses). Categories are defined from the task instruction: \emph{Playlists} operates on the user's Spotify playlists; \emph{Venmo friends} reads or edits the Venmo friend list; \emph{Venmo history} aggregates past Venmo payments or requests; \emph{Other} is every remaining task. Two tasks belong to two categories.}
  \label{tab:hierarchical-consolidation}
  {\footnotesize
  \setlength{\tabcolsep}{3.5pt}
  \begin{adjustbox}{max width=\linewidth}
  \begin{tabular}{l l cc cc cc >{\columncolor{gray!15}}c >{\columncolor{gray!15}}c cc}
    \toprule
    & & \multicolumn{2}{c}{Playlists (11)} & \multicolumn{2}{c}{Venmo friends (15)} & \multicolumn{2}{c}{Venmo history (13)} & \multicolumn{2}{>{\columncolor{gray!15}}c}{Other (131)} & \multicolumn{2}{c}{All (168)} \\
    \cmidrule(lr){3-4} \cmidrule(lr){5-6} \cmidrule(lr){7-8} \cmidrule(lr){9-10} \cmidrule(lr){11-12}
    Sessions & Consolidation & MSR & \passpowerfive & MSR & \passpowerfive & MSR & \passpowerfive & MSR & \passpowerfive & MSR & \passpowerfive \\
    \midrule
    \multicolumn{2}{l}{\emph{no memory}} & 45.5 & 9.1 & 42.7 & 0.0 & 32.3 & 0.0 & 46.0 & 18.3 & 44.3 & 14.9 \\
    \midrule
    90  & single call  & \textbf{67.3} & \textbf{45.5} & \textbf{69.3} & \textbf{20.0} & \textbf{55.4} & \textbf{7.7} & \textbf{59.2} & 34.4 & \textbf{60.2} & \textbf{32.1} \\
    140 & single call  & 49.1 & 27.3 & 36.0 & 6.7 & 52.3 & 0.0 & 56.2 & \textbf{36.6} & 53.7 & 30.9 \\
    140 & hierarchical & 41.8 & 0.0 & 57.3 & 13.3 & 43.1 & 0.0 & 56.8 & 30.5 & 54.6 & 25.0 \\
    \bottomrule
  \end{tabular}
  \end{adjustbox}}
\end{table}

\section{Memory Generation Settings}\label{sec:appendix-generation-process}

\paragraph{Setting the hyperparameters.}

\begin{figure}[h]
    \centering
    \begin{subfigure}[t]{0.47\textwidth}
        \includegraphics[width=\linewidth]{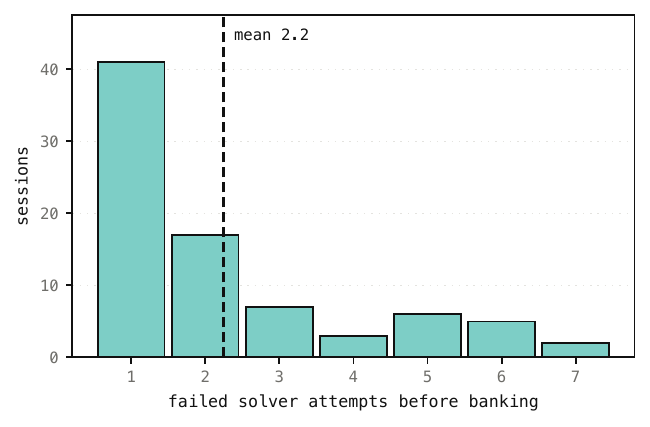}
        \caption{Failed solver attempts before $N_s=3$ consecutive successes.}
        \label{fig:num-failures}
    \end{subfigure}
    \hfill
    \begin{subfigure}[t]{0.47\textwidth}
        \includegraphics[width=\linewidth]{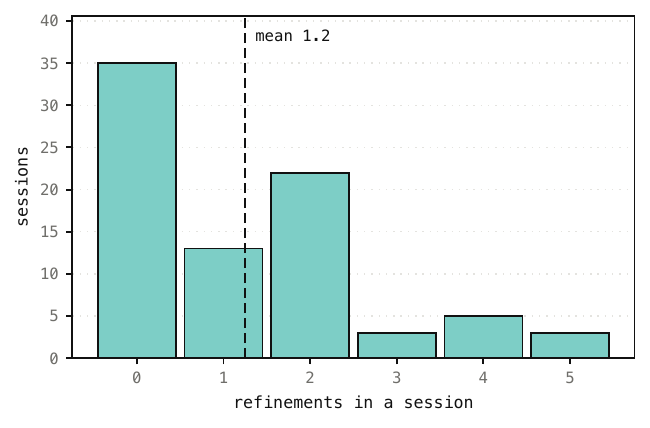}
        \caption{Task refinements in successful sessions.}
        \label{fig:num-refinements}
    \end{subfigure}
    \caption{Failures and refinements per accepted heuristic over the 90-session AppWorld run.}
    \label{fig:generation-parameters}
\end{figure}

The generation loop of \daedalus is controlled by a failure limit $N_f$, which caps the number of failed solver attempts at a task, and a number of refinements $N_r$ which sets the number of times the explorer may revise a task that is too easy or too hard. Low limits lose heuristics on the hardest tasks, which are the ones where memory helps most. High limits spend compute on tasks the Solver cannot solve or the Explorer cannot calibrate. We set $N_f=8$ and $N_r=5$ after a few preliminary sessions and validate them on the 90-session AppWorld run (Figure~\ref{fig:generation-parameters}). Most heuristics are accepted after one or two failures, but a tail of sessions need up to seven, so a smaller $N_f$ would lose heuristics on the hardest feasible tasks. Additionally, most sessions need at most two refinements and only a handful need four or five, which justifies our choice of $N_r=5$.

\paragraph{Impact of the extractor model choice.}\label{sec:apdx-extractor}
On the official AppWorld tasks, the benchmark provides the tasks and their verdicts, so the extractor is the only auxiliary agent that affects the memory. We vary its model and reasoning effort, while keeping GPT-5.4-mini as the solver (Table~\ref{tab:extractor-model}). With GPT-5.4-mini at high reasoning effort instead of GPT-5.4, the gain over the no-memory baseline falls from 16.5 to 12.8 MSR points. Disabling reasoning lowers it further to 9.5 points, while \passpowerfive does not change. Every extractor still improves the solver by a clear margin, so extraction from curated trajectories is robust to the choice of model. The drop from GPT-5.4 to GPT-5.4-mini (3.7 MSR points) is also smaller than on self-generated tasks (9.0 points, Section~\ref{sec:ablation-generation}), which suggests that the other auxiliary roles account for most of the drop in that setting.

\begin{table}[H]
  \centering
  \caption{Extractor model choice for \daedalus-curated on AppWorld.}
  \begin{tabular}{l l c c c}
    \toprule
    Extractor & Reasoning effort & \# Heuristics & MSR\up & \passpowerfive\up \\
    \midrule
    \emph{no memory} & -- & -- & 44.3\conf{1.1} & 14.9\confpass{2.8} \\
    \midrule
    GPT-5.4-mini & none & 26 & 53.8\conf{1.4} & 28.6\confpass{3.5} \\
    GPT-5.4-mini & high & 38 & 57.1\conf{1.1} & 28.6\confpass{3.5} \\
    GPT-5.4      & high & 40 & \textbf{60.8}\conf{0.9} & \textbf{36.3}\confpass{3.7} \\
    \bottomrule
  \end{tabular}
  \label{tab:extractor-model}
\end{table}

\paragraph{Counterfactual replay of accepted heuristics.}\label{sec:apdx-counterfactual}
Since the solver is an LLM agent, it is inherently stochastic and an ineffective heuristic could pass the acceptance rule by chance. To quantify to what extent the heuristics we accept raise the solver's success rate, we run the solver three times on each of the 81 accepted AppWorld tasks, with and without their accepted heuristic, and grade it with the LLM Judge. With its heuristic in context, the solver's MSR rises from 66.7\% to 77.0\% and its pass\textasciicircum{}3 from 40.7\% to 55.6\%. We assess these gains with a paired bootstrap, which resamples the 81 tasks with replacement 10{,}000 times while keeping each task's two conditions together, so that differences in difficulty between tasks do not inflate the variance. The resulting 95\% confidence intervals are [2.1, 18.5] points for MSR and [1.2, 28.4] points for pass\textasciicircum{}3, and both exclude zero.

\section{Memory Retrieval and Injection}
\label{app:retrieval-details}

\begin{figure}[H]
    \centering
    \includegraphics[width=0.95\linewidth]{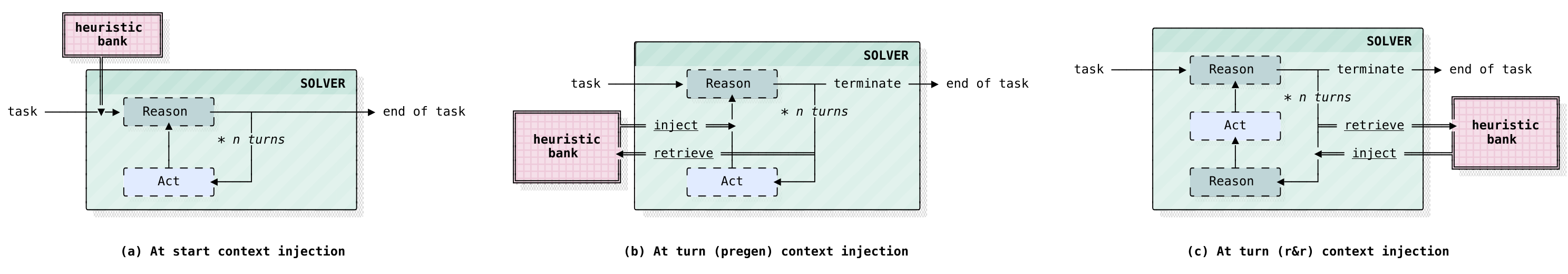}
    \caption{Heuristics injection modes at inference.}
    \label{fig:inference-modes}
\end{figure}

We investigate three axes that control how the heuristic bank is used at inference: the retrieval method and the number of retrieved heuristics, the source of the retrieval query, and the injection policy (Figure~\ref{fig:inference-modes}). All experiments use the 65-heuristic consolidated bank generated on AppWorld with GPT-5.4, evaluated on the 168 tasks of \texttt{test\_normal} over five runs with GPT-5.4-mini as the solver. Unless stated otherwise, retrieval uses the agent's reasoning from the previous turn as query (\textsc{Pre-Gen}), retrieves $k{=}5$ heuristics per turn, and keeps them in context with deduplication (\textsc{Cumulative, deduplicated}); this is the configuration reported in Table~\ref{tab:memory-use}.

\paragraph{The number of retrieved heuristics matters more than the retriever.}
At $k{=}5$, dense retrieval with Qwen3-Embedding-4B, sparse retrieval with BM25, and random selection do not differ significantly (Table~\ref{tab:memory-use}), indicating that on a consolidated bank, any five heuristics carry similar information regardless of how they are chosen. Increasing $k$ with BM25 improves both MSR and \passpowerfive{}, which plateau between $k{=}10$ and $k{=}20$ and rise again at $k{=}30$ (Figure~\ref{fig:topk-success}). Retrieval only approaches whole-bank injection once it surfaces about half of the bank per turn; for a bank of 3.1k tokens, retrieving a subset therefore brings no practical benefit.

\paragraph{Writing a dedicated retrieval query does not pay off.}
We compare two ways of forming the retrieval query (Table~\ref{tab:query-source}). \textsc{Pre-Gen} retrieves on the reasoning the agent produced at the previous turn, whereas \textsc{R2R} spends a second LLM call per turn to write a dedicated retrieval query. With BM25 in both cases, \textsc{R2R} more than triples the cost per run (14.8 vs.\ 3.8) and falls to the no-memory level on MSR (44.2 vs.\ 44.3) and below it on \passpowerfive{} (11.9 vs.\ 14.9). Writing a dedicated query does not make retrieval more precise and mostly adds noise, so we use \textsc{Pre-Gen} whenever retrieval is needed.

\paragraph{Heuristics help most when they persist in context.}
In Table~\ref{tab:injection-modes}, the retriever is fixed and only the way retrieved heuristics enter the context changes. \textsc{Transient} shows them for the current call and then drops them from the history. The other three policies keep them in the history and differ only in how they treat a heuristic that is retrieved again. \textsc{With repetition} appends it anyway, \textsc{Deduplicated} skips it, and \textsc{No replacement} removes it from the candidates altogether, so that each turn brings five heuristics the agent has not seen before. Retention is the dominant factor: \textsc{Transient} is the weakest policy (48.5 MSR), and its \passpowerfive{} of 13.1 falls below the no-memory baseline. Repeating heuristics already in context also hurts relative to deduplication (50.0 vs.\ 54.3 MSR), suggesting that redundant context harms multi-turn decision-making. \textsc{No replacement} exposes the solver to the most distinct heuristics (46.0 on average) and achieves the best retrieved-subset MSR (56.0), yet remains below whole-bank injection at task start (60.2). The same pattern holds for the whole bank: injecting it transiently at every turn (\textsc{All @ Turn}) underperforms injecting it once in the system prompt (\textsc{All @ Start}) on both metrics and at higher cost, consistent with the benefit of persistent guidance.

\begin{table}[t]
\centering
\small
\setlength{\tabcolsep}{6pt}
\caption{Retrieval query source on AppWorld. BM25, $k{=}5$, same consolidated bank.}
\label{tab:query-source}
\begin{tabular}{l c c r}
\toprule
Query source & MSR\up & \passpowerfive\up & \$\down \\
\midrule
No memory        & 44.3\conf{1.1} & 14.9\confpass{2.8} & \textbf{3.2} \\
\midrule
\textsc{Pre-Gen} & \textbf{52.0}\conf{1.3} & \textbf{22.0}\confpass{3.2} & 3.8 \\
\textsc{R2R}     & 44.2\conf{1.2} & 11.9\confpass{2.5} & 14.8 \\
\bottomrule
\end{tabular}
\end{table}

\begin{table}[t]
  \centering
  \small
  \setlength{\tabcolsep}{6pt}
  \caption{Injection modes on AppWorld. Retrieved subsets use Qwen3-Embedding-4B with \textsc{Pre-Gen} queries and $k{=}5$. Distinct: mean number of distinct heuristics seen by the solver at episode end. \textbf{Bold} marks the best value in each column and \underline{underline} the second best.}
  \label{tab:injection-modes}
  \begin{tabular}{l c c r c}
    \toprule
    Injection mode & MSR\up & \passpowerfive\up & \$\down & Distinct \\
    \midrule
    No memory & 44.3\conf{1.1} & 14.9\confpass{2.8} & 3.2 & 0 \\
    \midrule
    \rowcolor{gray!15}
    \multicolumn{5}{l}{\textit{Retrieved subset, $k{=}5$ per turn}} \\
    \textsc{Transient}                   & 48.5\conf{0.7} & 13.1\confpass{2.6} & \textbf{2.8} & 10.8 \\
    \textsc{Cumulative, with repetition} & 50.0\conf{1.7} & 17.3\confpass{2.9} & 3.1 & 11.7 \\
    \textsc{Cumulative, deduplicated}    & 54.3\conf{1.6} & 24.4\confpass{3.3} & 3.7 & 12.1 \\
    \textsc{Cumulative, no replacement}  & \underline{56.0}\conf{1.2} & 25.6\confpass{3.4} & 3.9 & 46.0 \\
    \midrule
    \rowcolor{gray!15}
    \multicolumn{5}{l}{\textit{Whole bank}} \\
    \textsc{All @ Turn} (transient) & 51.5\conf{1.2} & \underline{28.0}\confpass{3.5} & 3.5 & 65 \\
    \textsc{All @ Start}            & \textbf{60.2}\conf{0.9} & \textbf{32.1}\confpass{3.6} & 3.2 & 65 \\
    \bottomrule
  \end{tabular}
\end{table}

\section{Experimental Setup Details}\label{sec:apdx setup}

This section describes the configurations used to produce the main results (Table~\ref{tab:main-results}).

\paragraph{Models and shared settings.}
Every method uses the same pair of LLMs on a given benchmark (Table~\ref{tab:models}). The solver and the base agent run on the smaller model, and every auxiliary agent runs on the larger model. This includes roles that some original implementations assign to the solver model: the ExpeL reflector, the AutoGuide context module, the ReasoningBank extractor and judge, and the PREPING proposer, validator and curator. We raise them to the larger model so that no method is limited by a weaker auxiliary model. On $\tau^2$-bench, the user simulator is \texttt{gpt-4.1-2025-04-14} for all methods, as per the original benchmark implementation.

\begin{table}[h]
\centering
\small
\caption{Models and solver step ceiling for each benchmark (reasoning effort in parentheses).}
\label{tab:models}
\begin{tabular}{@{}llll@{}}
  \toprule
  \textbf{Benchmark} & \textbf{Solver / base agent} & \textbf{Auxiliary agents} & \textbf{Step ceiling} \\
  \midrule
  AppWorld             & \texttt{gpt-5.4-mini} (default) & \texttt{gpt-5.4} (high)       & 30 \\
  $\tau^2$-bench       & \texttt{gpt-5.4-mini} (default) & \texttt{gpt-5.4} (high)       & 200 \\
  AutomationBench      & \texttt{gpt-5.6-luna} (medium)  & \texttt{gpt-5.6-terra} (high) & 50 \\
  \bottomrule
\end{tabular}
\end{table}

\begin{wraptable}{r}{0.45\textwidth}
\centering
\small
\caption{\daedalus hyperparameters.}
\label{tab:hparams-daedalus}
\begin{tabular}{@{}lc@{}}
  \toprule
  \textbf{Hyperparameter} & \textbf{Value} \\
  \midrule
  Explorer turn ceiling                       & 40 \\
  Concurrent explorers                        & 5 \\
  Max task refinements $N_r$                  & 5 \\
  Max failed attempts $N_f$                   & 8 \\
  Consecutive successes to accept $N_s$       & 3 \\
  \bottomrule
\end{tabular}
\end{wraptable}
\paragraph{\daedalus.}
Table~\ref{tab:hparams-daedalus} lists the hyperparameters introduced in Algorithms~\ref{alg:main}--\ref{alg:solver}. \daedalus-curated uses the same values on the training tasks, with the benchmark verifier in place of the Judge. On the 90-session AppWorld run, an early experimental filter discarded seven accepted tasks whose expected tool paths near-duplicated banked tasks. We replaced them with seven successful sessions from the scaling experiment (Figure~\ref{fig:num-sessions}), yielding 88 accepted heuristics over the 90 retained sessions (97.8\%).  This filter was not part of the general method and was not used on other benchmarks. Analyses that use the tasks of this run (Section~\ref{sec:evaluation-proxies}, Figure~\ref{fig:generation-parameters}, and the counterfactual replay in Appendix~\ref{sec:apdx-counterfactual}) use only the 81 tasks accepted in the original sessions, without the seven replacements.

\paragraph{$\tau^2$-bench retail environment.}
In the retail domain, the tool \texttt{modify\_pending\_order\_items} assigns the price and options of the last exchanged pair to every item of a multi-item call, so two equally correct solutions that list the same items in a different order reach different database states and receive different grades. We do not modify the environment or the benchmark tasks, which keeps our scores comparable to the original implementation. Instead, the \daedalus{} Explorer rejects any generated task whose reference solution exchanges more than one item in a single call, so that generated tasks are not graded by argument order.

\paragraph{Baselines.}
We reimplement ExpeL, ERL, AutoGuide, ReasoningBank and ACE on top of our solver harness, with the hyperparameters of Table~\ref{tab:hparams-baselines}. For ExpeL, the insight list cap follows the authors' released code, and we use $L=3$ instead of the paper's 4 to 8 because our trajectories are much longer. For AutoGuide, we use $k=3$, the best value in the paper's own ablation. In ERL, we use both successful and failed attempts to create heuristics. ReasoningBank builds its bank on the training tasks, processed in parallel waves, and the bank is frozen at test time so we do not use test-time scaling (MaTTS). We use \texttt{text-embedding-3-large} as the embedding model to retrieve memories. For ACE, we use its offline variant without ground truth: each training task is solved once, sequentially, and a reflector and a curator turn every trajectory into ADD operations on the playbook, with no success signal. We start from an empty playbook, run the reflector and the curator on \texttt{gpt-5.4}, and inject the whole playbook at the start of each test task. For PREPING, we run the authors' released code and inject the whole playbook at the start of each test task. On $\tau^2$-bench, its task proposer also sees sampled database records, without which it could not write feasible tasks.
\begin{table}[h]
\centering
\small
\caption{Baseline hyperparameters.}
\label{tab:hparams-baselines}
\begin{tabular}{@{}llc@{}}
  \toprule
  \textbf{Method} & \textbf{Hyperparameter} & \textbf{Value} \\
  \midrule
  \multirow{6}{*}{ExpeL}
    & Max Reflexion retries per task $Z$            & 3 \\
    & Successes per comparison $L$                  & 3 \\
    & Max success/failure pairs per task            & 3 \\
    & Insight list cap                              & 20 \\
    & Few-shot examples $k$                         & 2 \\
    & Retrieval embedder                            & \texttt{all-mpnet-base-v2} \\
  \midrule
  ERL
    & Heuristics selected $k$                       & 20 \\
  \midrule
  \multirow{3}{*}{AutoGuide}
    & Max Reflexion retries per task                & 3 \\
    & Max success/failure pairs per task            & 3 \\
    & Guidelines injected $k$                       & 3 \\
  \midrule
  \multirow{4}{*}{ReasoningBank}
    & Memory items per trajectory                   & $\leq 3$ \\
    & Experiences retrieved $k$                     & 1 \\
    & Retrieval embedder                            & \texttt{text-embedding-3-large} \\
    & Tasks per wave                                & 5 / 5 / 8 \\
  \midrule
  \multirow{7}{*}{ACE}
    & Rollouts per training task                    & 1 \\
    & Tasks per wave                                & 1 \\
    & Curator operations                            & ADD only \\
    & Initial playbook                              & empty (8 sections) \\
    & Playbook token budget                         & 80k \\
    & Final playbook bullets                        & 240 / 164 / 103 \\
  \midrule
  \multirow{3}{*}{PREPING}
    & Cycles $\times$ tasks per cycle               & $10 \times 10$ \\
    & Feasibility score to update the playbook      & 5 \\
    & Completion score to count as success          & $\geq 4$ \\
  \bottomrule
\end{tabular}
\end{table}

\paragraph{Memory construction budget.}
\daedalus{} spends more solver rollouts than any baseline, about 6 times more than ExpeL and 14 times more than single-attempt methods (Table~\ref{tab:rollouts}). Most of this gap is the cost of not having training tasks. Of the 1{,}225 rollouts, 520 are spent on the explorer's first version of each task, which is close to the 592 rollouts of \daedalus{}-curated on the training tasks, and the remaining 705 are spent on refined versions of tasks that were too easy or too hard. The Solver loop itself therefore costs about as much on generated tasks as on curated ones, and task calibration roughly doubles it. These rollouts use the small Solver model, so they account for a minor share of the generation cost reported in Table~\ref{tab:pipeline-ablation}, which is dominated by the explorer and the judge.

\paragraph{Environment access during generation.}
\daedalus{} never interacts with the environment states of test tasks. On AppWorld, each generation session runs in the initial world of one training task, whose instruction and verifier are not used, and generation excludes Amazon and Gmail, which appear only in the challenge split~\citep{trivedi2024appworld}. The Explorer therefore sees the same environment states as methods using training tasks. On AutomationBench, generation uses only the worlds of the 30 training tasks. On $\tau^2$-bench, all tasks share a single database, which methods using training tasks also access.
\begin{table}[h]
\centering
\small
\caption{Solver rollouts spent building memory on AppWorld. Each \daedalus{} session counts as one training task. Explorer interactions are not included.}
\label{tab:rollouts}
\begin{tabular}{l c c c}
  \toprule
  Method & Solver rollouts & Per task & MSR \\
  \midrule
  ERL                  & 90        & 1.0  & 58.6 \\
  ReasoningBank        & 90        & 1.0  & 48.0 \\
  ACE                  & 90        & 1.0  & 60.5 \\
  ExpeL                & 200       & 2.2  & 59.0 \\
  AutoGuide            & 200       & 2.2  & 44.0 \\
  \daedalus{}-curated  & 592       & 6.6  & 60.8 \\
  \daedalus{}          & 1{,}225   & 13.6 & 60.2 \\
  \bottomrule
\end{tabular}
\end{table}

\section{Generation Examples on AppWorld} \label{sec:examples}

\subsection{Generated Task and Accepted Heuristic}

Table~\ref{tab:examples} shows one task generated by the Explorer on AppWorld, the success conditions the Judge checks, and the heuristic accepted after the Solver loop. Section~\ref{sec:example-session} details a full generation session.

\begin{table}[H]
\centering
\caption{AppWorld example}
\label{tab:examples}
\begin{tabular}{p{0.1\linewidth} p{0.8\linewidth}}
  \toprule
  Task \text{example} & Please text everyone from my latest friends dinner note to let them know how much they owe me for that dinner. \\
  \midrule
  Success \text{conditions} & - Compared with the starting state, every person listed in the latest friends dinner note has a new outgoing text message from Kristin in that contact's message thread. \newline - Each new text correctly states the amount assigned to that person in the note and makes clear it is about that dinner.\\
  \midrule
  Accepted heuristic & - When using \texttt{simple\_note}, the available note-reading methods are \texttt{search\_notes} and \texttt{show\_note}; there is no \texttt{list\_notes}, so search by query first and then open the chosen \texttt{note\_id}. \newline - When API-doc search results are suppressed, print \texttt{dir(apis.<app>)} to discover exact method names and proceed from those concrete names instead of guessing endpoints. \newline - When a cross-app workflow needs another app after notes, get and use that app's own access token; tokens are app-specific and a token from one app will 401 on another app's endpoints. \newline - When a login with the main profile email fails for the phone app, try authenticating with the user's phone number as the username, because that app may use phone-number-based login rather than email. \newline - When a search query returns noisy matches, do not trust the query alone; inspect the returned titles and timestamps and open the candidate note to confirm it is the latest relevant one before acting on its contents. \\
  \bottomrule
\end{tabular}
\end{table}

\subsection{A Complete Generation session} \label{sec:example-session}

\begin{figure}[H]
    \centering
    \includegraphics[width=\linewidth]{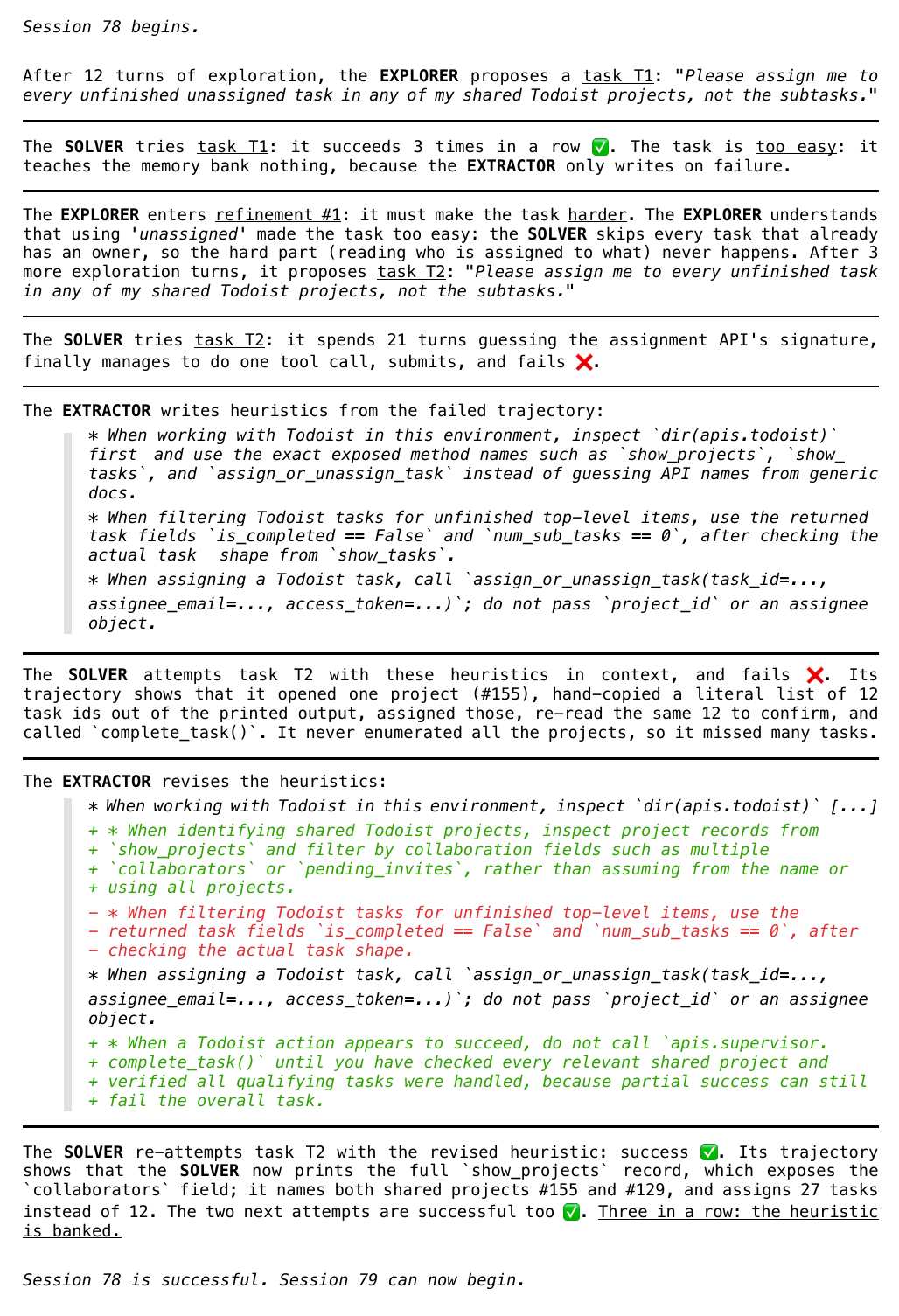}
    \caption{Session 78 from the \daedalus generation process on AppWorld (partial).}
    \label{fig:example-session}
\end{figure}

\subsection{Surveyor Output}

\begin{textbox}[AppWorld's Survey Output]{gen}
\pline{\textbf{Axis} — The useful spread here is by app/domain involvement, not by raw endpoint count: Spotify, Todoist, Splitwise, Venmo, Phone, File System, Simple Note, plus the tiny meta areas api\_docs and supervisor. That partition matches the real records and errands I actually saw: Spotify has 324 searchable songs over 17 pages plus playlists, queue state and premium receipts; Todoist has shared projects/sections/tasks/subtasks/comments/attachments; Splitwise has groups, expenses, payments and balances; Venmo has transactions, requests, balance and comments; Phone has 133 texts over 7 pages plus contacts and alarms; File System and Simple Note are much narrower CRUD helpers; api\_docs and supervisor are mostly scaffolding. Because many good tasks span two apps, treat these as involvement buckets rather than forcing every task into one exclusive owner.}
\pskip
\pline{\textbf{Distribution} — Weighted, with most of the run in the four rich end-user domains and only a small cap on meta/shallow work. Spotify — 22\% — broad, non-interchangeable music work: library, playlists, queue/player, likes/downloads, and receipt export to file\_system all behaved differently. Todoist — 18\% — real planning/collaboration work is dense even with only 4 projects listed; \texttt{delete\_\allowbreak{}section} removed task \texttt{6334} and subtask \texttt{9636}, and invites/comments/attachments add distinct task shapes. Splitwise — 17\% — fewer surfaces than Spotify but high-value reconciliation work: expense \texttt{5040} and payment \texttt{6015} changed balances immediately, and \texttt{delete\_\allowbreak{}group 274} cascaded to associated expenses/payments.}
\pelide
\pskip
\pline{\textbf{Guidance for the task generator} — A well-covered run should lean hard into Spotify, Todoist, Splitwise and Venmo, with Phone used regularly and File System often as a helper in cross-app tasks, not just as isolated mkdir/cat boilerplate.}
\pelide
\pskip
\pline{\textbf{Pitfalls} — The easiest over-coverage trap is shallow CRUD in Simple Note, File System, api\_docs and supervisor: yes, they work, but too many ``create a note'', ``make a folder'', or ``show the active task'' requests would miss where the environment's real complexity lives. The easiest under-coverage trap is Splitwise and Todoist, whose value exceeds their surface count: balances changed after every expense/payment write, \texttt{delete\_\allowbreak{}group} soft-deleted related records, and \texttt{delete\_\allowbreak{}section} erased its tasks and subtasks.}
\pelide
\end{textbox}

\FloatBarrier

\section{Prompts}\label{sec:prompts}

Each prompt combines a template shared by all benchmarks (Appendix~\ref{sec:prompts-daedalus}) with benchmark-specific blocks, such as a description of the environment and the protocol for acting in it. We reproduce the shared templates, the AppWorld solver blocks, and the two benchmark-specific Extractor prompts (Appendix~\ref{sec:prompts-benchmark}); all prompts are available in full in the GitHub repository. Prompts marked \texttt{(partial)} are truncated, with [\ldots] marking an elided passage. Values filled in at run time appear as \pph{placeholders}, and [Identical to Prompt N] marks text shared with another prompt. Boxes are colored by role: task generation ({\color{pGenFrame}\textbf{blue}}), judging ({\color{pJudFrame}\textbf{red}}), memory extraction and consolidation ({\color{pMemFrame}\textbf{green}}), and solving ({\color{pSolFrame}\textbf{yellow}}).

\subsection{\daedalus{} Prompts}\label{sec:prompts-daedalus}

\begin{promptbox}[Surveyor (shared template, partial)]{gen}\label{prompt:surveyor}
\pline{You are an environment analyst. A task-generation system will soon invent many tasks in this environment (a solver attempts them, so reusable know-how can be mined from the attempts). Your job is not to design tasks. Your job is to work out, from the environment itself, what a good spread of those tasks looks like — the coverage goal — and to justify it from what you observe.}
\pskip
\phead{The question to answer}
\pline{Over a whole generation run the tasks should not pile onto one comfortable corner. But ``spread out'' only means something once you choose the axis to spread along and the distribution over it. Both are yours to decide:}
\pitem{1}{\textbullet}{The axis might be a sub-area or module, a group of related operations, a capability or operation type, a data domain, or a task shape (a question answered from the data, a sweep over everything that matches, a stateful change, a multi-area reconciliation), or a combination. Choose whatever genuinely partitions the useful work here, and say why even coverage of the obvious partition would or would not be right.}
\pitem{1}{\textbullet}{The distribution can be even or weighted. \textbf{Do not weight by how many operations an area exposes.} Endpoint count measures how much API was written, not how much real work lives there: an area with sixty operations may be sixty variations on one shallow idea, while an area with twelve may carry most of what a person actually does here. Weight instead by how much of a real person's genuine use of this environment falls in each area, and by how much distinct, non-interchangeable work each supports. Give a concrete emphasis and a one-line reason per unit, and if you find yourself ranking areas in the same order as their operation counts, stop and re-justify it.}
\pelide
\pline{\textbf{You may not deliver the report until you can answer all five of these for EVERY area, from something you actually ran} — not from its documentation, and not by analogy with another area:}
\pskip
\pitem{1}{1.}{What does a real record here contain, and which of its fields connect it to another area?}
\pitem{1}{2.}{What is past the first page of its main listing — how many pages are there, and does the listing behave differently beyond page one?}
\pitem{1}{3.}{What happens when you pass a plausible but wrong identifier, and what happens when you pass a stale one from an earlier call?}
\pitem{1}{4.}{What does one of its write operations do to \emph{related} records, checked by reading them back afterwards?}
\pitem{1}{5.}{What does it refuse to do, and with what error — a policy limit, an eligibility rule, or an already-in-that-state no-op?}
\pelide
\phead{Output, part 1: the coverage goal}
\pline{Four short subsections, in this order, each under its own \texttt{**heading**} and written as a few plain sentences:}
\pskip
\pline{\textbf{Axis} — the axis (or combination) you chose, and why it is the right partition here, with the evidence.}
\pline{\textbf{Distribution} — even or weighted, and the emphasis for each unit along the axis with a short reason, grounded in realistic use rather than endpoint count. This is the one place where a compact list helps: one line per unit, \texttt{unit — share — reason}.}
\pline{\textbf{Guidance for the task generator} — what to emphasize, what to cap, how to vary the kind of request, and what a well-covered run looks like.}
\pline{\textbf{Pitfalls} — what is easy to over-cover (shallow or boilerplate areas), what is easy to under-cover (areas whose real value exceeds their surface count), and the traps a naive generator would miss (exact ids, pagination, completeness, cross-area steps).}
\pelide
\phead{Output, part 2: the tag manifest}
\pline{The report on its own cannot be measured, so also define a closed set of tags spanning the axis you chose. Each tag is a short, reusable label a designer assigns to the task they just built. Each carries:}
\pitem{1}{\textbullet}{\texttt{tag}: a short, stable, '\#'-prefixed label, unique in the set;}
\pitem{1}{\textbullet}{\texttt{definition}: one line stating when a designer should assign it, concrete enough that two designers would agree;}
\pitem{1}{\textbullet}{\texttt{target\_\allowbreak{}share}: a number in (0, 1], the fraction of banked tasks that should carry the tag.}
\pskip
\pline{Three rules decide whether a tag is worth defining, and they matter more than how many you write:}
\pskip
\pline{\textbf{A tag must be able to come out false.} If nearly every task you would want here carries it, it measures nothing and steers nothing — a designer sees it satisfied and moves on. Anything you would give a target much above half is a baseline requirement, not a coverage axis: say it in \textbf{Guidance} instead and leave it out of the manifest. Prefer tags whose target share sits in the range where being under it is a real, visible gap.}
\pelide
\pline{\pph{benchmark-specific deliverable-partition rule, if any}}
\pelide
\pline{\pph{benchmark-specific survey protocol}}
\end{promptbox}

\begin{promptbox}[Explorer (shared template, partial)]{gen}\label{prompt:explorer}
\pline{You are a task designer exploring a live environment. Invent ONE task that is genuinely solvable HERE, confirm it is solvable, and hand off a clean specification for a separate solver agent to attempt later.}
\pskip
\pline{\pph{benchmark-specific task guidelines}}
\pskip
\phead{What makes a good task}
\pitem{1}{\textbullet}{Grounded: it references data that actually exists here (real records, entities, values you discovered) — never invented ones. Inspect the data first to find the specifics.}
\pelide
\pitem{1}{\textbullet}{Natural: it reads like a real request a person would actually make — a plain, purposeful errand or question, not a puzzle. Difficulty must come from the genuine effort of the request and from navigating real policy/eligibility rules — NOT from contrived constraints a person would never state. Strictly avoid invented metrics, multi-level tie-breaker chains, embedded lists, and any instruction a human would never spell out.}
\pitem{1}{\textbullet}{Under-specified on purpose: leave OUT every value the solver could discover for itself. Naming a value you had to go and find deletes the work you meant to set. State a literal only when the person supplies it from outside the environment.}
\pitem{1}{\textbullet}{Hard enough, in the right way: calibrate against a CAPABLE solver — if a competent agent would very likely finish it on the first try, it is too easy, and an easy task is wasted effort for the whole run. But get that difficulty from SCOPE, not from size: from what the request leaves the solver to decide (which records qualify, how far back to look, whether there is more than one page, what to do when several match or none do), and from real policy or eligibility limits. Do NOT reach for difficulty by chaining more steps, touching more parts of the environment, or demanding more side effects — that makes a task longer, not harder. But a SHORT request is not a SMALL job: keep the wording brief and the work substantial — one clause covering many records or several pages. A task the solver finishes in a handful of calls is too small however naturally it reads.}
\pelide
\pskip
\pline{Tasks already generated (do something DIFFERENT):}
\pitem{1}{\textbullet}{\pph{previously accepted task 1}}
\pitem{1}{\textbullet}{\pph{previously accepted task 2}}
\pitem{1}{\textbullet}{...}
\pskip
\phead{Coverage goal (from the one-time environment survey)}
\pline{A read-only survey of this environment decided how the whole run's tasks should be spread. Treat it as the coverage plan you are executing:}
\pline{\pph{coverage-goal report produced by the Surveyor}}
\pskip
\phead{Coverage so far — steer toward the UNDER tags}
\pline{\pph{live tally of coverage tags over the banked tasks, marked UNDER/OK/OVER}}
\pskip
\phead{Guidelines from past generations}
\pline{Follow these closely — they encode what made previous tasks too easy or too hard:}
\pitem{1}{\textbullet}{\pph{evolved guideline 1}}
\pitem{1}{\textbullet}{\pph{evolved guideline 2}}
\pitem{1}{\textbullet}{...}
\pelide
\phead{The specification you produce}
\pline{However you emit it (see the protocol below), your specification has three parts:}
\pitem{1}{\textbullet}{the task: a short, natural-language request a person would actually make. Self-contained in the sense that nothing OUTSIDE the environment is missing — not in the sense of listing what is inside it. Before you emit it, reread it once and delete every id, address, path, amount, date, count and record name that the solver could have found by looking; refer to those things the way the person would. Then check it states a goal rather than a sequence of steps;}
\pitem{1}{\textbullet}{the minimal solution: the ordered tool calls that would solve it (not your exploration trail);}
\pitem{1}{\textbullet}{success conditions: outcome-based, path-agnostic checks on the END STATE, each expressed as a change from the starting state. Never reference which tools were used or how. Only require what the task actually determines — do not bake in incidental details (exact free-text wording, ids, timestamps) the solver could not have known, unless the task itself dictates them; allow any reasonable phrasing for free text.}
\pskip
\pline{\pph{benchmark-specific environment block}}
\pskip
\pline{\pph{benchmark-specific explorer protocol}}
\end{promptbox}

\begin{promptbox}[Explorer task-design doctrine (shared, partial)]{gen}\label{prompt:doctrine}
\pline{\textbf{Write a request, not a specification.} One to three short sentences. A person asking for something does not enumerate the steps, does not spell out the edge cases, and does not paste in record ids. If your instruction has grown into a paragraph with semicolons and a step-by-step plan, it has stopped being a request.}
\pskip
\pline{\textbf{State the goal; never the procedure.} Say what \pph{subject} wants to be true afterwards, or what they want to know. Do not sequence the work with ``then'', ``after that'', ``next'', ``finally''. Do not say which \pph{route\_noun} or which call to use for each part. Working out the route IS the task — if you hand over the route, the \pph{actor} only has to type.}
\pskip
\pline{\textbf{Short to state is not small to do.} Keep the words brief and the work substantial: \pph{example}. Shorten the wording, never the amount of \pph{swept\_noun} that has to be \pph{verb}. A request the}
\pline{\pph{actor} finishes in a handful of calls is too small however well it reads.}
\pskip
\pline{Say scope the way people do — a superlative (\pph{superlative}), a universal (\pph{universal}), or a relative time frame (\pph{timeframe}). Each is short to write and expensive to get right, so treat them as your main source of difficulty, not as tricks to avoid.}
\pelide
\end{promptbox}

\begin{promptbox}[Explorer, refinement of an off-target task (shared template)]{gen}\label{prompt:explorer-refine}
\pline{A task you designed earlier turned out **\pph{too\_easy | too\_hard}** when a solver attempted it. Revise it so it lands closer to the right difficulty — challenging but solvable — while keeping it the SAME task in spirit (same goal and style). Make it slightly \pph{harder | easier} with a small, targeted change, not a redesign; keep it grounded and natural, and re-confirm it is solvable before you emit it (produce the same specification under the same rules as your original design).}
\pskip
\pline{\pph{benchmark-specific task guidelines}}
\pskip
\pline{Those rules bind the revision too. In particular, when a task came back \textbf{too\_hard} the fix is NOT to name the records, spell out the steps, or narrow it to a listed subset — that deletes the work instead of easing it. Make it easier by reducing how much has to be swept or reconciled, or by removing one requirement, while the request stays short and still makes the solver find things for itself.}
\pskip
\pline{The task as it stands:}
\pline{\pph{current task specification}}
\pskip
\pline{Revisions already tried this session (do NOT repeat these):}
\pitem{1}{\textbullet}{[\pph{too\_easy | too\_hard}] \pph{earlier variant}}
\pskip
\pline{Your earlier exploration (reuse what you already learned — you need not rediscover it):}
\pline{\pph{transcript of the Explorer's earlier exploration}}
\pskip
\pline{\pph{benchmark-specific refinement protocol}}
\end{promptbox}

\begin{promptbox}[Explorer guideline update (partial)]{gen}\label{prompt:explorer-guidelines}
\pline{You maintain a list of guidelines for a task designer. The designer invents tasks in a live environment that are then attempted by a separate solver agent. A task is only useful if it is \textbf{challenging but solvable} — neither trivial (solved on the first attempt) nor impossible (never solved) — AND it must read like a natural request a real person would make, never a contrived puzzle. Your guidelines steer the designer toward that frontier WITHOUT sacrificing naturalness. Difficulty should come from genuine work (cross-referencing real data across tools, applications, or data areas, several real steps), never from artificial rules — so do NOT write guidelines that push invented metrics, multi-level tie-breakers, embedded lists, or other things a human would never say. Concrete, environment-specific guidelines are welcome when the evidence supports them.}
\pskip
\pelide
\pskip
\pline{Current guidelines:}
\pitem{1}{\textbullet}{\pph{evolved guideline 1}}
\pitem{1}{\textbullet}{\pph{evolved guideline 2}}
\pitem{1}{\textbullet}{...}
\pskip
\pline{All tasks so far that were TOO EASY (solved on the first attempt):}
\pitem{1}{\textbullet}{\pph{too-easy task}}
\pitem{1}{\textbullet}{...}
\pskip
\pline{All tasks so far that were TOO HARD (never solved):}
\pitem{1}{\textbullet}{\pph{too-hard task}}
\pitem{1}{\textbullet}{...}
\pskip
\pline{This session's trajectory — the designer proposed a task, the solver attempted it, and (if it was off-target) the designer revised it up to several times. Each entry shows the task variant, how it turned out (too\_easy / too\_hard / banked = good difficulty), and the solver's trace. Study the PROGRESSION to learn what kind of change moved the difficulty in the right direction (or failed to):}
\pskip
\pline{--- variant 1: \pph{too\_easy | too\_hard | banked} ---}
\pline{TASK: \pph{task variant}}
\pline{SOLVER TRACE:}
\pline{\pph{solver trace}}
\pskip
\pline{Update the guideline list using the evidence above — especially how the difficulty changed across the variants. You may ADD a new guideline, MODIFY an existing one, or DELETE one that the evidence now contradicts. Keep guidelines that still hold.}
\pelide
\pline{Maintain a COMPACT, curated playbook of DISTINCT, general design principles — it is the only difficulty signal the task designer sees, so keep it sharp. This is a curated set, NOT a running log: every guideline must earn its place, and the list must not grow without bound. Cover what makes a task too easy here, what makes one too hard, and the concrete design levers in between that raise difficulty \emph{naturally}. Never suggest artificial difficulty. Each guideline is a GENERAL principle — concrete and specific about task \emph{design} and \emph{difficulty}, never referencing specific entities or values — not a session-specific observation.}
\pelide
\pline{Output ONLY the consolidated list — one guideline per line, most important first, no numbering, no prose, no markdown.}
\end{promptbox}

\begin{promptbox}[LLM judge (partial)]{judge}\label{prompt:judge}
\prole{System}
\pline{You verify whether an agent achieved a task's intended outcomes. Judge the end effect, not exhaustive proof: a successful tool call (optionally confirmed by a read) is enough evidence its effect happened. Fail only on clear evidence a required outcome was missed or wrong; give competent solutions the benefit of the doubt and don't penalize incidental details you can't see.}
\pskip
\prole{User}
\pline{You are evaluating whether an agent completed a task. Judge by the EFFECTS visible in its trajectory — not by the agent's own claims, and not by demanding exhaustive proof of every detail.}
\pskip
\pline{TASK:}
\pline{\pph{task instruction}}
\pskip
\pline{The task counts as COMPLETE only if all of these intended outcomes hold at the end:}
\pskip
\pitem{1}{1.}{\pph{success condition 1}}
\pitem{1}{2.}{\pph{success condition 2}}
\pitem{1}{3.}{...}
\pskip
\pline{AGENT TRAJECTORY (the actions it took — code, tool calls, and/or messages — and the observed results):}
\pskip
\pline{\pph{solver trajectory}}
\pskip
\pline{How to judge, outcome by outcome:}
\pelide
\pitem{1}{\textbullet}{Trust effects, not narration. A tool call that returns a success message (optionally confirmed by a follow-up read) is sufficient evidence its effect happened — you do NOT need every field re-printed. Conversely, the agent saying ``done'' or otherwise signaling completion does NOT by itself prove anything.}
\pitem{1}{\textbullet}{An outcome is UNMET if the trajectory shows it was skipped, applied to the wrong target or value, left an unresolved error, or only partially done.}
\pelide
\pitem{1}{\textbullet}{An outcome covering a GROUP (``every'', ``each'', ``all the ... that ...'') is met only on evidence about the whole group, not about the records the agent happened to handle. A listing can come back partial and look complete, so N actions on one listing's N results say nothing about the group's size. If the trajectory gives you no way to establish that size, that outcome is UNMET — the benefit of the doubt above covers ambiguous detail, never completeness.}
\pskip
\pline{The task SUCCEEDS only if every intended outcome is met; otherwise it FAILS.}
\pskip
\pline{Output STRICTLY one JSON object on a single line, no prose, no code fences:}
\texttt{\{"success": true or false, "reason": "\textless{}one short sentence naming the deciding factor\textgreater{}"\}}
\end{promptbox}

\begin{promptbox}[Extractor (shared; used as is on AppWorld; partial)]{mem}\label{prompt:extractor}
\prole{System}
\pline{You help an agent build reusable memory from task attempts. Given a trajectory and outcome, produce or refine a single memory item that captures actionable lessons for future attempts. Output ONLY the memory text — no headers, no JSON, no fences.}
\pskip
\prole{User}
\pline{You maintain a running memory for a specific task the agent is retrying. The memory is a bullet list of discrete, actionable lessons that is injected into the agent's prompt on its next attempt. The agent has just made an attempt; produce the updated list.}
\pskip
\pline{Task: \pph{task instruction}}
\pline{Attempt: \#\pph{attempt index} (failure limit: \pph{max failures})}
\pline{Outcome: \pph{success | failure}}
\pskip
\pline{Current memory (from previous attempts):}
\pline{\pph{current heuristic, if any}}
\pskip
\pline{Agent trajectory:}
\pline{\pph{solver trajectory}}
\pskip
\phead{How to update the list}
\pline{If a memory already exists, rewrite it rather than appending: copy correct bullets unchanged, edit ones that need refining, drop ones this attempt proved wrong, and add bullets for what was just learned. Prefer a few sharp bullets over many vague ones.}
\pskip
\pelide
\pskip
\pline{Look across all the failure dimensions, not only execution errors:}
\pitem{1}{\textbullet}{Task comprehension: if the agent acted without reading all of the task's context (e.g. replied before reading the message that prompted the request, or acted on everything instead of the filtered subset the task names), add a bullet on what to read first and why.}
\pitem{1}{\textbullet}{Answer submission: if the task wants a factual answer (a count, title, name, yes/no) and the agent worked it out but never returned it in the form the environment expects, add a bullet on returning the answer correctly.}
\pitem{1}{\textbullet}{Efficiency: if the agent did more work than needed — e.g. retrieved far more than necessary and filtered afterwards when the environment can filter or query directly — add a bullet recommending the targeted call.}
\pskip
\pelide
\pskip
\phead{Output}
\pline{Write each bullet as a single line beginning with \texttt{- }:}
\pitem{1}{\textbullet}{When \textless{}situation\textgreater{}, \textless{}do this\textgreater{}[, because \textless{}the reason\textgreater{}]. Keep every bullet self-contained and generalizable — no task-specific identifiers (user ids, emails, values) and nothing that must be derived while solving (e.g. ``the count is X''). Output only the bullet list, starting directly with \texttt{- }. No headers, JSON, fences, or commentary.}
\end{promptbox}

\begin{promptbox}[Consolidation prompt (partial)]{mem}\label{prompt:consolidation}
\pline{You are an expert at distilling an AI agent's scattered experience into a compact, high-signal playbook. The agent solves tasks in an interactive environment by issuing tool / API calls (in some environments, by writing and executing code that calls them). Below are \pph{N} reflection notes the agent wrote after solving individual tasks. They are extremely REDUNDANT (the same few lessons restated many times) and polluted with task-specific detail (``for this task, first look up record X, then …''). Your job: turn them into a clean playbook of ATOMIC, CONCRETE, GENERALIZABLE heuristics. Rules for the output heuristics:}
\pitem{1}{1.}{ATOMIC — each heuristic states exactly ONE lesson, in a sentence or two. No multi-lesson bullets.}
\pitem{1}{2.}{DEDUPLICATE — when notes clearly restate the SAME lesson, emit it once in its sharpest form; if you are unsure whether two notes make the same point, keep both.}
\pitem{1}{3.}{CONCRETE — preserve exact, fully-qualified tool / API / function names and parameter names EXACTLY as the notes give them, in whatever call syntax the environment uses. Concreteness is the whole point; never abstract a real call into vague advice.}
\pitem{1}{4.}{GENERALIZABLE — state reusable mechanisms, not recipes welded to one task. DROP every ``for this task / for this workflow'' instruction and anything that only makes sense for one specific task's data.}
\pelide
\pline{Return STRICT JSON only (no markdown fence), an object:}
\texttt{\{"heuristics": [\{"topic": "\textless{}short topic tag for the sub-area, e.g. general | auth | discovery | output | completion | or an environment-specific tool/area name\textgreater{}", "text": "\textless{}the atomic heuristic\textgreater{}"\}, ...]\} Output the smallest set of heuristics that still fully covers the material — let the notes decide the count; never aim for a target number. Two goals in tension:}
\pitem{3}{1.}{COVERAGE — nothing useful is lost, so every distinct reusable lesson the notes contain is represented, spanning both general mechanisms that apply to every task and environment-specific facts about the particular tools or areas the notes cover;}
\pitem{3}{2.}{MINIMALITY — no redundancy, each distinct lesson stated exactly once in its sharpest form, merging overlapping notes rather than listing variants (do not pad the list).}
\pline{Here are the notes:}
\pline{\pph{numbered list of the N retained heuristics}}
\end{promptbox}

\begin{promptbox}[Solver (shared template)]{solver}\label{prompt:solver}
\pline{You are an autonomous agent that completes a user's task in a live environment.}
\pskip
\pline{Work carefully and deliberately:}
\pitem{1}{\textbullet}{Read the task and identify every condition, filter, and constraint it states, and make sure your solution satisfies all of them.}
\pitem{1}{\textbullet}{Inspect the real data before you act — never assume a value or an identifier; look it up in the environment.}
\pitem{1}{\textbullet}{Proceed one step at a time: after each action, check the result before deciding the next step.}
\pitem{1}{\textbullet}{Handle errors gracefully — if an action fails, read the error and adjust rather than repeating it.}
\pitem{1}{\textbullet}{Verify that the intended outcome actually holds before you finish, and finish only once the task is genuinely complete.}
\pskip
\pline{\pph{benchmark-specific environment block}}
\pskip
\pline{\pph{benchmark-specific solver protocol}}
\pskip
\pline{Learned guidelines from previous experience. These guidelines are advisory and may not apply to the current task. Use only those that are consistent with the task, the environment policy, the available tools, and the data you observe.}
\pskip
\pitem{1}{1.}{\pph{heuristic 1}}
\pitem{1}{2.}{\pph{heuristic 2}}
\pitem{1}{3.}{...}
\end{promptbox}

\subsection{Benchmark-specific prompts}\label{sec:prompts-benchmark}

The original AppWorld solver prompt~\citep{trivedi2024appworld} contains extensive human-written heuristics for the environment, so we replace it with a simplified version (Prompt~\ref{prompt:app-world-new}, inserted into the shared solver template of Prompt~\ref{prompt:solver}) to evaluate all methods with less hand-engineered guidance.

\begin{promptbox}[AppWorld solver used in this work (blocks of Prompt~\ref{prompt:solver})]{solver}\label{prompt:app-world-new}
\prole{Environment block}
\phead{What you already know about this environment}
\pitem{1}{\textbullet}{Most app APIs are protected. Get credentials with \texttt{apis.\allowbreak{}supervisor.\allowbreak{}show\_\allowbreak{}account\_\allowbreak{}passwords()} (a list of \texttt{\{account\_\allowbreak{}name,\allowbreak{} password\}}) and pick the entry matching the app you need.}
\pitem{1}{\textbullet}{Log in through the app's own login method (e.g. \texttt{apis.\allowbreak{}\textless{}app\textgreater{}.\allowbreak{}login(username=...\allowbreak{},\allowbreak{} password=...\allowbreak{})}). The username is usually the main user's profile email, not their display name.}
\pitem{1}{\textbullet}{Do not assume a login session persists: capture the returned \texttt{access\_\allowbreak{}token} and pass it explicitly (\texttt{access\_\allowbreak{}token=...\allowbreak{}}) to every later protected call for that app.}
\pitem{1}{\textbullet}{Before filtering or aggregating records, inspect one returned item's fields so you filter on the real field name rather than guessing.}
\pskip
\prole{Protocol}
\pline{You act by writing and executing Python code in a persistent, interactive environment (variables persist between steps). Each turn, respond in exactly this format:}
\pskip
\pline{Thought: \textless{}your reasoning about what to do next\textgreater{}}
\pskip
\texttt{\textasciigrave{}\textasciigrave{}\textasciigrave{}python\\ \textless{}your code\textgreater{}\\ \textasciigrave{}\textasciigrave{}\textasciigrave{}}\par
\pskip
\pitem{1}{\textbullet}{Call application APIs directly as \texttt{apis.\allowbreak{}\textless{}app\_\allowbreak{}name\textgreater{}.\allowbreak{}\textless{}method\_\allowbreak{}name\textgreater{}(...\allowbreak{})}.}
\pitem{1}{\textbullet}{Discover APIs with \texttt{apis.\allowbreak{}api\_\allowbreak{}docs.\allowbreak{}search\_\allowbreak{}api\_\allowbreak{}docs(query="...\allowbreak{}")} (free-form text) or \texttt{dir(apis.\allowbreak{}\textless{}app\_\allowbreak{}name\textgreater{})}, then \texttt{apis.\allowbreak{}api\_\allowbreak{}docs.\allowbreak{}show\_\allowbreak{}api\_\allowbreak{}doc(app\_\allowbreak{}name="...\allowbreak{}",\allowbreak{} api\_\allowbreak{}name="...\allowbreak{}")} for the full documentation.}
\pitem{1}{\textbullet}{When the task is complete, call \texttt{apis.\allowbreak{}supervisor.\allowbreak{}complete\_\allowbreak{}task()}. If and only if the task asks for a factual answer (a number, name, title, yes/no, ...), pass it as \texttt{apis.\allowbreak{}supervisor.\allowbreak{}complete\_\allowbreak{}task(answer=\textless{}your\_\allowbreak{}answer\textgreater{})} where the answer is a concise string or number. If the task only requires side effects (sending a message, saving a file, following an artist, ...), call it with no arguments.}
\pitem{1}{\textbullet}{Make \texttt{complete\_\allowbreak{}task(...\allowbreak{})} the ONLY statement in its code block, and submit it only after an earlier turn has verified the result (re-read the records you changed, re-derive any answer from returned data). Do not compute or fix anything in the same block as \texttt{complete\_\allowbreak{}task}.}
\pskip
\pline{The applications available to you (these are the ONLY ones — treat any other app as nonexistent):}
\pline{\pph{JSON list of app names and descriptions}}
\pskip
\pline{The main user you are helping is: \pph{supervisor's name, email and phone number}. Authenticate as needed.}
\end{promptbox}

\begin{promptbox}[$\tau^2$-bench Extractor (partial)]{mem}\label{prompt:tau2-extractor}
\prole{System}
\pnote{Identical to Prompt \ref{prompt:extractor}}
\pskip
\prole{User}
\pline{You maintain a running memory for a specific customer-service task the agent is retrying. The memory is a bullet list of discrete, actionable lessons that is injected into the agent's prompt on its next attempt. The agent has just made an attempt; produce the updated list.}
\pskip
\pnote{Identical to Prompt \ref{prompt:extractor}}
\pskip
\pline{Agent trajectory (conversation + tool calls):}
\pline{\pph{solver trajectory}}
\pskip
\pnote{Identical to Prompt \ref{prompt:extractor}}
\pskip
\pelide
\pskip
\pline{Look across all the failure dimensions, not only tool errors:}
\pitem{1}{\textbullet}{Policy compliance: if the agent broke a policy rule (skipped a required eligibility check, acted without the mandatory explicit confirmation before a modify/cancel/refund, or exceeded what policy allows), add a bullet stating the rule and when it applies.}
\pitem{1}{\textbullet}{Information gathering: if the agent acted before authenticating the user or before looking up the records the action depends on, add a bullet on what to establish first and why.}
\pitem{1}{\textbullet}{Tool arguments: if a call failed or silently did the wrong thing from a wrong or missing argument (wrong id, wrong enum value, a list where one item was meant), add a bullet with the correct usage.}
\pitem{1}{\textbullet}{Communication: if the task needed the agent to tell the user something (a total, a status, an eligibility answer) and it never did, add a bullet on reporting results before ending.}
\pitem{1}{\textbullet}{Ending: if the agent stopped or transferred early, or kept going after the request was complete, add a bullet on when the task is actually done.}
\pskip
\phead{Output}
\pline{Write each bullet as a single line beginning with \texttt{- }:}
\pitem{1}{\textbullet}{When \textless{}situation\textgreater{}, \textless{}do this\textgreater{}[, because \textless{}the reason\textgreater{}]. Keep every bullet self-contained and generalizable — no task-specific identifiers (user ids, order numbers, names, addresses) and nothing that must be derived while solving (e.g. a specific order total). Output only the bullet list, starting directly with \texttt{- }. No headers, JSON, fences, or commentary.}
\end{promptbox}

\begin{promptbox}[AutomationBench Extractor (partial)]{mem}\label{prompt:automationbench-extractor}
\prole{System}
\pnote{Identical to Prompt \ref{prompt:extractor}}
\pskip
\prole{User}
\pline{You maintain a running memory for a specific business-operations task the agent is retrying. The task runs against simulated SaaS applications (mail, chat, spreadsheets, calendars, trackers, knowledge bases, CRM, e-signature) reached through a single HTTP tool. The memory is a bullet list of discrete, actionable lessons that is injected into the agent's prompt on its next attempt. The agent has just made an attempt; produce the updated list.}
\pskip
\pnote{Identical to Prompt \ref{prompt:extractor}}
\pskip
\phead{What makes a bullet worth writing}
\pline{The agent will next meet a DIFFERENT task, in a DIFFERENT world, wired to a DIFFERENT set of applications. A bullet earns its place only if it would still change what the agent does there.}
\pskip
\pline{Write decision rules over the task's own data — which records to read before acting, which ones qualify, which source wins when two disagree, what must be true before a write, what to check after one. These recur across tasks because the reasoning recurs.}
\pskip
\pline{Do not write a recipe for one endpoint. A bullet whose situation clause is ``to create a ticket in system X'' or ``to lock a page in system Y'' fires only when the next task happens to use that call, and the agent can already discover the call itself. The exception is a mechanical trap that bites across services — argument encoding, placeholder substitution in a discovered URL, what a response does and does not prove — which is worth one bullet stated once, in general terms.}
\pskip
\pline{Look across all the failure dimensions, not only the calls that returned an error:}
\pitem{1}{\textbullet}{Evidence gathering: if the agent selected, ranked or excluded records before reading every source that governs the choice (all tabs of a worksheet, the full thread rather than its first message, the channel history rather than the newest post), add a bullet on what to read first and why.}
\pitem{1}{\textbullet}{Qualification: if the agent applied the task's filters, exclusions, holds or date scopes loosely, or applied a requirement to records that never declared it, add a bullet stating how to decide membership.}
\pitem{1}{\textbullet}{Authority: if the agent treated the wrong source as decisive — an older notice over a newer one, an unofficial mention over the authorized sender, a request field over the registered record — add a bullet naming which source governs.}
\pelide
\pitem{1}{\textbullet}{Write verification: if the agent reported completion on the strength of a call that returned without error, add a bullet on what to re-read to prove the effect landed.}
\pelide
\phead{Output}
\pline{Write each bullet as a single line beginning with \texttt{- }:}
\pitem{1}{\textbullet}{When \textless{}situation\textgreater{}, \textless{}do this\textgreater{}[, because \textless{}the reason\textgreater{}]. State the situation by the shape of the task, not by the application it happened to use. Keep every bullet self-contained and generalizable — no task-specific identifiers (record ids, addresses, spreadsheet or channel names, values) and nothing that must be derived while solving (e.g. ``the count is X''). Output only the bullet list, starting directly with \texttt{- }. No headers, JSON, fences, or commentary.}
\end{promptbox}

\end{document}

%% file: math_commands.tex
\usepackage{amsmath,amsfonts,bm}

\def\eqref#1{equation~\ref{#1}}

\def\1{\bm{1}}

\DeclareMathAlphabet{\mathsfit}{\encodingdefault}{\sfdefault}{m}{sl}
\SetMathAlphabet{\mathsfit}{bold}{\encodingdefault}{\sfdefault}{bx}{n}

